\documentclass{article}
\usepackage{arxiv}
\usepackage[utf8]{inputenc} 
\usepackage[T1]{fontenc}    
\usepackage{hyperref}       
\usepackage{amsmath,amssymb,amsfonts}
\usepackage{url}            
\usepackage{booktabs}       
\usepackage{amsfonts}       
\usepackage{nicefrac}       
\usepackage{microtype}      
\usepackage{cleveref}       
\usepackage{lipsum}         
\usepackage{graphicx}
\usepackage{xcolor}
\usepackage{appendix}
\usepackage{subcaption}
\usepackage{multicol}
\usepackage{doi}
\usepackage{comment}
\usepackage[ruled,vlined]{algorithm2e}
\usepackage{xcolor}

\SetCommentSty{mycommfont}
\usepackage[backend=bibtex, isbn=false, url=false, sorting=none]{biblatex}
\title{Explaining Human Activity Recognition with SHAP: Validating Insights with Perturbation and Quantitative Measures}

\date{\today}

\newif\ifuniqueAffiliation
\uniqueAffiliationtrue

\ifuniqueAffiliation 
\author{ \href{https://orcid.org/0009-0005-6310-408X}{\includegraphics[scale=0.06]{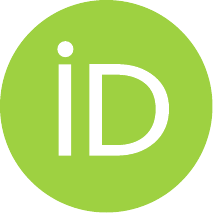}\hspace{1mm}Felix Tempel}\\
	Faculty of Informatics\\
	Norwegian University of Science and Technology\\
	Trondheim, Norway \\
	\texttt{felix.e.f.tempel@ntnu.no} \\
 	\And
    \href{https://orcid.org/0000-0002-2469-1809}{\includegraphics[scale=0.06]{orcid.pdf}\hspace{1mm}Espen Alexander F. Ihlen} \\
	Faculty of Medicine and Health Sciences\\
	Norwegian University of Science and Technology\\
	Trondheim, Norway \\
	\texttt{espen.ihlen@ntnu.no} \\
	\And
	\href{https://orcid.org/0000-0001-5532-0034}{\includegraphics[scale=0.06]{orcid.pdf}\hspace{1mm}Lars Adde} \\ 
	Department of Clinical and Molecular Medicine\\
	Norwegian University of Science and Technology\\
    Clinic of Rehabilitation \\
    St. Olavs Hospital, Trondheim University Hospital\\
	Trondheim, Norway \\
	\texttt{lars.adde@ntnu.no} \\
    \And
	\href{https://orcid.org/0000-0003-1820-6544}{\includegraphics[scale=0.06]{orcid.pdf}\hspace{1mm}Inga Strümke} \\
	Faculty of Informatics\\
	Norwegian University of Science and Technology\\
	Trondheim, Norway \\
	\texttt{inga.strumke@ntnu.no} \\
}

\renewcommand{\headeright}{}

\usepackage[colorinlistoftodos,prependcaption,textsize=tiny]{todonotes}

\hypersetup{
pdfauthor={Felix Tempel},
}

\begin{document}
\maketitle

\begin{abstract}
In Human Activity Recognition (HAR), understanding the intricacy of body movements within high-risk applications is essential. 
This study uses SHapley Additive exPlanations (SHAP) to explain the decision-making process of Graph Convolution Networks (GCNs) when classifying activities with skeleton data.
We employ SHAP to explain two real-world datasets: one for cerebral palsy (CP) classification and the widely used NTU RGB+D 60 action recognition dataset.
To test the explanation, we introduce a novel perturbation approach that modifies the model's edge importance matrix, allowing us to evaluate the impact of specific body key points on prediction outcomes. 
To assess the fidelity of our explanations, we employ informed perturbation, targeting body key points identified as important by SHAP and comparing them against random perturbation as a control condition.
This perturbation enables a judgment on whether the body key points are truly influential or non-influential based on the SHAP values.
Results on both datasets show that body key points identified as important through SHAP have the largest influence on the accuracy, specificity, and sensitivity metrics. 
Our findings highlight that SHAP can provide granular insights into the input feature contribution to the prediction outcome of GCNs in HAR tasks. 
This demonstrates the potential for more interpretable and trustworthy models in high-stakes applications like healthcare or rehabilitation. 
\end{abstract}

\section{Introduction}
\label{sec:introduction}

In recent years, there has been a notable expansion in the field of developing Graph Convolution Networks (GCNs) for Human Activity Recognition (HAR).
GCNs have established themselves as the default method for classifying human activities based on skeleton data \cite{shuchangSurveyHumanAction2022}.
Nevertheless, the principal objective of researchers has been to enhance the utilized GCN architectures, particularly on benchmark datasets such as NTU RGB+D 60/120 \cite{shahroudyNTURGB+DLarge2016, liuNTURGB+D1202020}, and Kinetics \cite{kayKineticsHumanAction2017}. 
Because of this, exploring the underlying mechanisms of the developed GCN models has been largely neglected. 
At the same time, these models' increasing complexity and size have led to a lack of interpretability, effectively turning them into black boxes.
Consequently, there is a conspicuous research gap between examining the accuracy and interpretability of GCNs in the HAR domain, limiting the applicability of these methods in high-risk fields such as medical and clinical decision-making \cite{bharatiReviewExplainableArtificial2024, chaddadSurveyExplainableAI2023}.

To address this issue, methods within the field of Explainable Artificial Intelligence (XAI) could be employed to shed light on the internal mechanisms of GCNs, thereby fostering trust and providing informed decision-making foundations for the users of such systems.
This is particularly crucial within the medical domain and other high-risk areas, where understanding the reasoning behind AI-generated outcomes is vital for users to make informed decisions \cite{bharatiReviewExplainableArtificial2024, aliExplainableArtificialIntelligence2023, nazirSurveyExplainableArtificial2023, tjoaSurveyExplainableArtificial2021}.
Therefore, meaningful explanations provided by - or alongside the predictions of - GCN models must be transparent and comprehensible for the users to accept and safely use such systems within their respective environment \cite{chaddadSurveyExplainableAI2023, saeedExplainableAIXAI2023}.

Most XAI methods for GCNs rely on gradient-based explanation concepts to interpret their models \cite{dasGradientWeightedClassActivation2022, Song2022ConstructingSA, pellanoMovementsMetricsEvaluating2024}.
These methods are currently preferred due to their direct correlation with the skeleton body key points, allowing the identification of the specific body parts important for the model's prediction.
However, while such approaches effectively highlight body key points, they fail to capture the nuanced contributions of other essential input features - such as position, velocity, acceleration, and segment orientation - vital for a deeper understanding.

SHapley Additive exPlanations (SHAP) \cite{lundbergUnifiedApproachInterpreting2017} can overcome this fundamental shortcoming of gradient-based methods as it provides a direct relationship between the different input features and model predictions, offering a more comprehensive understanding of how each input feature contributes to the final prediction of the GCNs.
This level of detailed insight is not possible with gradient-based methods, which typically provide only a prevailing sense of the general importance of the features by probing the overall gradient at the last network layer.
Conversely, SHAP assigns an exact contribution value to each feature for a given prediction, facilitating a more granular and interpretable understanding of how each feature influences the model's output.
Although SHAP exhibits substantial potential compared to gradient-based methods, it has yet to be implemented in the context of GCNs within HAR.
This gap presents an opportunity to enhance the accuracy and granularity of model interpretation. 
The comprehensive explanations provided by SHAP have the potential to make it an effective method for interpreting HAR models.

While XAI plays a pivotal role in making machine learning (ML) more understandable, trustworthy, and accessible, commonly used methods often rely heavily on visual, qualitative validation methods, which is also the case for the few endeavors within the HAR domain. 
In many cases, researchers tend to cherry-pick results that support their stated claim, emphasizing the critical need for comprehensive and standardized evaluation frameworks for XAI methodologies \cite{leavittFalsifiableInterpretabilityResearch2020, millerExplanationArtificialIntelligence2019}.
This reliance highlights the need for quantitative metrics to objectively evaluate explanation methods beyond what visual and qualitative assessments alone can deliver \cite{nautaAnecdotalEvidenceQuantitative2023}.
Since single performance metrics such as a high accuracy score do not provide insights into how and why an ML model makes its predictions, the interpretability of a complex ML system cannot be reduced to one simple metric, necessitating a multifaceted perspective instead \cite{nautaAnecdotalEvidenceQuantitative2023, doshi-velezRigorousScienceInterpretable2017, molnarInterpretableMachineLearning2022}.

A prominent method for evaluating XAI methods is the perturbation-based approach \cite{fisherAllModelsAre2019, ivanovsPerturbationbasedMethodsExplaining2021}.
In essence, by observing how sensitive a model's predictions are to changes in particular inputs, one can gain insight into which features are most important or influential for decision-making.
This method allows for an investigation of how modifications to input features impact the model's predictions, thus enabling an examination of the model's explanation.
This technique is powerful due to its intuitive nature and broad applicability across various ML models, rendering it a straightforward and effective technique for XAI testing.
Despite its potential, this approach has not yet been explored within the HAR domain, and there is a conspicuous absence of perturbation methods that provide relevant insights.
Within HAR, where understanding the model's reasoning process can be crucial for interpreting predictions related to physical motion, the lack of tailored perturbation methods suggests an opportunity for further research and development. 
Extending perturbation techniques to HAR at the model level, where the model is treated as a dynamic entity rather than a static, unchangeable component, could yield more meaningful insights from explanation methods such as SHAP. 
Such an approach can enhance the transparency and reliability of the model's decision-making processes.

\subsection{Contributions}
In this paper, we present several novel contributions addressing the gaps mentioned earlier within current HAR and XAI research:

\begin{itemize}
    \item First work applying Shapley value-based explanation to a GCN model for human skeleton data on the primary input features, encapsulated within a new algorithmic framework named ShapGCN. 
    \item Introduction of a novel perturbation approach tailored to GCNs within the domain of HAR to test the explanation.
    \item Apply quantitative metrics to validate the explanations with the perturbation on two real-world HAR datasets.
\end{itemize}

These contributions collectively advance the understanding of HAR by bridging the gap between model performance and interpretability, offering valuable insights for researchers and practitioners in the respective HAR domain.

\section{Related Work}

\subsection{XAI Methods}
Several XAI approaches have been proposed to explain model predictions within the healthcare and HAR domain since modern ML systems require explainability tools to ensure transparency and trust \cite{bharatiReviewExplainableArtificial2024, chaddadSurveyExplainableAI2023, tjoaSurveyExplainableArtificial2021, allgaierHowDoesModel2023, sadeghiReviewExplainableArtificial2024}.
Among the most common are SHAP \cite{lundbergUnifiedApproachInterpreting2017} used in our study, saliency maps \cite{simonyanDeepConvolutionalNetworks2014}, Local Interpretable Model-agnostic Explanations (LIME) \cite{ribeiroWhyShouldTrust2016}, Gradient-weighted Class Activation Mapping (GradCAM) \cite{selvarajuGradCAMVisualExplanations2020}, and Concept Activation Vectors (CAVs) \cite{kimInterpretabilityFeatureAttribution2018}.

Saliency maps are used to visualize the region of interest of an input attribution based on the model's decision \cite{simonyanDeepConvolutionalNetworks2014}.
Saliency maps can be applied to visualize the importance of the individual input channels, which can then be used for dimensionality reduction and improved computational efficiency as done in \cite{yanResNetLikeCNNArchitecture2022}.
However, saliency maps are limited in that they typically only provide a global and sometimes unstable representation of the model's decision process \cite{adebayoSanityChecksSaliency2018}.

LIME is a popular model-agnostic approach for interpretability, providing model explanations by use of interpretable surrogate models. This is achieved by perturbing points from the input dataset in a neighborhood in order to generate training data points used to fit an interpretable model in this region. 
This interpretable model is used as a directly interpretable surrogate of the full model around the explained data point \cite{ribeiroWhyShouldTrust2016}.
While LIME is effective across a variety of models, it has not been applied within the HAR domain to the best of our knowledge. 
This may be due to the challenge of generating meaningful perturbations in skeleton data, as perturbations to this kind of data may not translate to realistic kinematic sound human movement, making the generated explanations less reliable.

GradCAM is a gradient-based method providing visual explanations by highlighting regions of the input based on the gradient activation from a chosen layer from the model \cite{selvarajuGradCAMVisualExplanations2020}.
However, GradCAM focuses on the global relevance of the gradients taken from the model's chosen layer without precisely quantifying the individual input feature contributions to the output.

Another XAI approach is the use of CAVs, which explain model predictions in terms of high-level human-understandable concepts \cite{kimInterpretabilityFeatureAttribution2018}.
This method can be applied to domains where well-defined concepts exist, such as in medical imaging, where features like the presence of a tumor can be interpreted as a meaningful concept for a model's decision \cite{borysExplainableAIMedical2023, lucieriInterpretabilityDeepLearning2020}.
However, applying CAVs to skeleton HAR poses significant challenges.
Constructing well-defined and human-intuitive concept functions for human activities in skeleton data, which often involve complex and subtle movements across multiple body key points, would likely be excessively involved and time-consuming. 
As such, concept-based explanations have only been applied to accelerometer sensor data within the HAR domain \cite{jeyakumarXCHARConceptbasedExplainable2023}.

\subsection{XAI within HAR}

Within the domain of skeleton-based HAR, most XAI approaches are mostly limited to gradient-based explanation techniques like class activation maps (CAM) \cite{zhouLearningDeepFeatures2016} or GradCAM \cite{selvarajuGradCAMVisualExplanations2020}. 
In contrast, the widely used SHAP \cite{lundbergUnifiedApproachInterpreting2017}, a method based on cooperative game theory that provides consistent and locally accurate feature attributions, has, to the best of our knowledge, only been employed in one work within the HAR domain although only on the secondary input features.

One relevant work that adapts the gradient-based explanation concept is from \cite{dasGradientWeightedClassActivation2022}.
The authors use GradCAM to localize and visually highlight the important body key points within the skeleton data. 
Additionally, the validity is tested by masking and perturbing parts of the skeleton to calculate metrics, including faithfulness and contrastivity \cite{popeExplainabilityMethodsGraph2019}.
However, their gradient approach misses the potential to reveal the individual contribution of the input features to the model output, which other model-agnostic methods like SHAP could achieve.
Moreover, their perturbation approach reduces the overall accuracy, regardless of the importance of the specific joint being masked, which is the key parameter used in their metrics.
An alternative approach to the perturbation, which would not entirely occlude the skeleton body key points from the model, could potentially provide more realistic and kinematic sound explanation results.

Another work using a gradient-based approach is \cite{Song2022ConstructingSA}, where the authors explore the utilization of CAM mapped to the skeleton sequences. 
Their results only offer visual representations of the derived explanations without incorporating contextual analysis or evaluative measures.
Consequently, in the absence of objective metrics to assess the explanation produced, the overall contribution of this approach remains challenging to compare to other concepts.
As with the work from \cite{dasGradientWeightedClassActivation2022}, this gradient approach misses the potential to reveal the individual contribution of the features to the final output.
This is particularly important since \cite{Song2022ConstructingSA} uses multiple branches to train their network on individual input representations derived from the skeleton data fused later in the model.
Again, approaches like SHAP could investigate the importance of those individual input features more deeply.

In \cite{pellanoMovementsMetricsEvaluating2024}, the authors also use a gradient-based approach to examine how well XAI techniques work for HAR by assessing the metrics Stability and Faithfulness, first proposed in \cite{agarwalOpenXAITransparentEvaluation2024} and \cite{agarwalRethinkingStabilityAttributionbased2022}, on GradCAM and CAM values. 
Their work aims to simulate realistic changes in input data by applying perturbations to the NTU RGB+D 60 dataset.
The EfficientGCN \cite{Song2022ConstructingSA} architecture is used as the model architecture for testing the generated GradCAM and CAM values through the used metrics.
However, the metrics for the Prediction Gap of important (PGI) and the Prediction Gap of unimportant (PGU) features show the same trend as randomly perturbing body key points.
This indicates that the perturbation method applied does not demonstrate any significant advantage over random perturbation, thereby questioning its effectiveness in distinguishing between important and unimportant body key points.
Since higher values of the stability metrics from \cite{agarwalRethinkingStabilityAttributionbased2022}, indicate higher instability of the explanation, a logical conclusion would be that no perturbation would result in the lowest value.
However, the trend observed in \cite{pellanoMovementsMetricsEvaluating2024} reveals a distinct pattern, showing that the highest values occur with minor perturbation.
This contrasts the metric discussed in \cite{agarwalRethinkingStabilityAttributionbased2022}, again causing us to question the perturbation procedure.
A different approach for the perturbation, which also encodes the spatial information of the skeleton, could lead to more realistic explanation results.
Since this work uses the architecture from \cite{Song2022ConstructingSA}, the importance of the individual feature contribution also holds up here.

In \cite{gaoAutomatingGeneralMovements2023}, the authors use SHAP to enhance the model's explanations for early cerebral palsy (CP) screening.
Their work employs three branches, which combine infant videos and essential characteristical inputs to classify CP.
However, the explainability aspect is restricted to the secondary input features (sex, gestational age, birth weight, and the corrected age), limiting its overall impact on the model's interpretability.
This narrow focus on secondary input features restricts the generalizability of the findings and overlooks significant insights that could be derived from the primary features in the video.
Expanding the focus to include primary input features could fully leverage SHAP's capabilities to improve the model's transparency and interpretability.
Overall, it can be stated that SHAP has some advantages compared to gradient-based explanations in depicting the particular contribution of individual features.

\section{ShapGCN}

\begin{figure}
    \centering
    \includegraphics[scale=1.2]{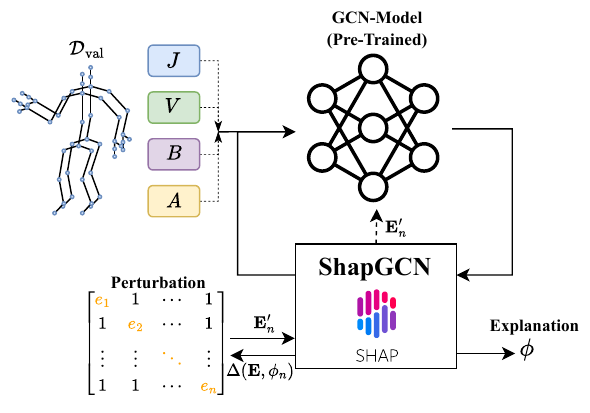}
    \caption{\textbf{Overview of the proposed XAI pipeline for HAR - ShapGCN.} 
    The features computed from the skeleton data are fed into a pre-trained model, followed by the explanation through ShapGCN to determine the feature attributions on the primary input features. 
    The explanation $\phi$ is validated with the proposed quantitative metrics $\mathcal{M}$ and interpreted visually. 
    Based on $\phi$, the perturbed edge matrix $\mathbf{E}^{\prime}_n$ is constructed, and the model architecture is updated.
    Afterward, $\mathcal{D}_{val}$ is inferred on the updated model architecture.}
    \label{fig:shapgcn}
\end{figure}

This section provides the essential technical background to follow our contributions. 
We begin with an overview of SHAP and its application to GCNs in HAR. 
Next, we present the novel perturbation approach developed to evaluate and validate the explanations generated by our framework ShapGCN. 
We conclude by describing the metrics used to evaluate the obtained explanations.
In Fig.~\ref{fig:shapgcn}, the general workflow of the proposed ShapGCN algorithm is shown.

\subsection{SHAP}
SHAP is a unified framework for explaining ML model predictions by attributing an importance score to each model input feature \cite{lundbergUnifiedApproachInterpreting2017}.
This is done by adapting a solution concept from game theory, the Shapley value \cite{shapley7ValueNPerson1953}, to the context of a predictive ML model by formulating the model input features and prediction in the game theoretic setting, thus creating the SHAP value.
Representing the importance of a model input feature ${i}$, the SHAP value $\phi$ is computed as follows:
\begin{equation}
    \phi_i = \sum_{S \subseteq F \setminus \{i\}} \frac{|S|!(|F|-|S|-1)!}{|F|!} [f_{S \cup \{i\}}(x_{S \cup \{i\}}) - f_{S}(x_{S})].
\end{equation}
To estimate the Shapley value, the model's input features must be systematically removed from the model during prediction. 
This process requires evaluating the model on all input feature subsets $S \subseteq F$, where $F$ represents the set of all features and results in importance scores for each feature, denoting their impact on the model's prediction.
This impact is computed by comparing predictions when the feature is included ($f_{S \cup \{i\}}$) versus excluded ($f_S$). 
These differences are computed across all the subsets of $S \subseteq F \setminus \{i\}$, and the final SHAP values are derived as a weighted average of these differences. 
Since a trained ML model cannot be evaluated with input features missing, SHAP simulates the absence of input features by replacing their values with a mean or uninformative value.
The SHAP algorithm generates an attribution score for each model input feature, denoted as $\phi_i(x)$, quantifying its contribution to the model’s output. 
These SHAP values break down the model’s prediction into contributions from each feature.
The sum of these contributions for all features approximates the difference between the reference value $\phi_0$ and the model’s prediction $f(x)$ for a given input $x$, which can be written as:

\begin{equation}
    f(x) \approx \sum_{i=1}^{N} \phi_i(x) + \phi_0,
\end{equation}

where each $\phi_i(x)$ quantifies how much the specific feature drives the model prediction away from the reference value $\phi_0$ for the data instance $x$.
Here, the reference value, $\phi_0$, is the expected output of the model, calculated as the average over the entire input distribution:

\begin{equation}
    \label{eq:phi_shap}
    \phi_0 = \mathbb{E}_x[f(x)].
\end{equation}

In our approach, the input features are computed from the skeleton data, divided into four distinct groups, namely: \textbf{1) Position ($J$)}, \textbf{2) Velocity} ($V$), \textbf{3) Bone ($B$)}, and \textbf{4) Acceleration ($A$)} features, as defined in \cite{tempelAutoGCNgenericHumanActivity2024, tempelLightweightNeuralArchitecture2024}. 
\textbf{1) $J$}: These are the 3D or 2D coordinates of each body key point, capturing the spatial layout of the skeleton.
\textbf{2) $V$}: Velocity measures how fast each body key point moves between frames, calculated as the difference in position over time, to capture the dynamics of the motion.
\textbf{3) $B$}: These features represent the relative positions between connected body key points, encoding the length and angular orientation to capture the spatial relationships during motion.
\textbf{4) $A$}: Acceleration captures the rate of change of velocity, highlighting sudden changes in motion like starts, stops, or rapid shifts in direction.
Aggregating the SHAP values of each of these feature groups, we calculate the total SHAP value as $\phi = \phi_J + \phi_V + \phi_B + \phi_A$.
The data used to calculate $\phi_0$ in our approach is randomly sampled from $\mathcal{D}_{train}$ with $n=100$.

The computation of exact SHAP values has a time complexity of $\mathcal{O}(2^n)$, where $n$ is the number of input features, making it computationally expensive for high-dimensional datasets. 
However, several approximation methods have been developed to make SHAP more applicable for real-world usage. 
These include algorithms such as ``Kernel SHAP'' for model-agnostic interpretation, ``Tree SHAP'' for tree-based models, and ``Deep SHAP'', which utilizes the ``Deep Lift'' algorithm for deep learning models \cite{lundbergUnifiedApproachInterpreting2017}. 
These approximations significantly reduce the complexity while maintaining the interpretability and consistency of SHAP values.
Beyond explaining individual predictions or local explanations, SHAP values can be aggregated across multiple data points to get global explanations \cite{chaddadSurveyExplainableAI2023}. 
By summing the SHAP values for each feature across the dataset, we obtain a global understanding of the most influential features overall. 
This provides insight into how the model makes decisions on a broader scale, allowing us to identify the most critical features driving predictions across the dataset.

\subsection{Perturbation for GCNs}
In XAI, perturbation techniques are commonly employed to perform sensitivity analyses of a model on its input features by perturbing portions of the input data and observing the resulting impact on the model's predictions \cite{xiongExplainableArtificialIntelligence2024}. 
In our novel perturbation approach, we extend this technique by perturbing the parts of the model architecture directly correlating with the skeleton body key points instead of perturbing the input features.
In the following, we give a concise overview of the GCN architecture utilized in skeleton HAR before elaborating further on our methodology.

GCNs extend the concept of convolutional neural networks to graph data, making them suitable for skeleton data. 
In a GCN, each node aggregates features from its neighboring nodes, effectively capturing local and spatial relationships inherent to human movements. 
The core operation involves multiplying the node feature matrix by a normalized adjacency matrix $\mathbf{A}$, identity matrix $\mathbf{I}$, and a learnable edge importance weight matrix $\mathbf{E}$ to capture the spatial dependencies of the skeleton.
Depending on the chosen partition strategy $s = \{\text{spatial}, \text{distance}, \text{uniform}\}$, $\mathbf{A}$ is parsed into $\mathbf{A}_j$, where $\mathbf{A} + \mathbf{I} = \sum_j \mathbf{A}_j$.
Based on $s$, the maximum graph distance $j$ defines the output as follows \cite{yanSpatialTemporalGraph2018}:

\begin{equation}
	\label{eq:graphconv}
	\mathbf{f}_{out} =  \sum_{j} \mathbf{W}_{j} \mathbf{f}_{in} (\mathbf{\Lambda}_{j}^{-\frac{1}{2}} \mathbf{A}_{j} \odot \mathbf{E}_{j} \mathbf{\Lambda}_{j}^{-\frac{1}{2}}) \,.
\end{equation}

Here, $\mathbf{f}_{out}$ denotes the output features and $\mathbf{f}_{in}$ the input features respectively. 
$\mathbf{W}_{j}$, $\mathbf{\Lambda}_{j}$, and $\mathbf{A}_{j}$ contribute to the convolutional operation, normalization of $\mathbf{A}$, and the adjacency matrix itself with graph distance $j$.
Matrix $\mathbf{E}$ has the same shape as $\mathbf{A}$, and its elements $e$ are the learnable parameters along the diagonal, defined as

\[
\mathbf{E} = \begin{bmatrix}
    \color{orange}{e_{1}} & 1 & \cdots   & 1 \\
    1 & \color{orange}{e_{2}} & \cdots   & 1 \\
    \vdots & \vdots & \color{orange}{\ddots}  & \vdots \\
    1 & 1 & \cdots  & \color{orange}{e_{n}}
\end{bmatrix}.
\]

The diagonal elements $e_n$ are emphasized in orange, indicating their learnable nature during training, while the off-diagonal elements are fixed to 1.

Given that this matrix encodes spatial information, it represents the significance of edges in different architecture regions to the model.
This opens the possibility of applying an informed perturbation to the model by masking the respective edges corresponding to a skeletal body key point.
The individual importance of those edges at a specific time is obtained through the SHAP values; see Fig.~\ref{fig:shapgcn} for an overview of the overall procedure.
In our experimental setting, we assigned a value of $e_n = 1 \times 10^{-5}$ to the respective edges, simulating a masking effect. 
Alternatively, we applied a percentage-based perturbation to introduce variation, modifying the original edge values as $e_n = e_n \times \epsilon$, where $\epsilon$ represents the percentage adjustment.

\subsection{XAI Metrics}
\label{sec:metrics}
The validity of the explanations obtained from SHAP is tested through the presented perturbation procedure.
To assess it, the PGI and PGU metrics are utilized \cite{agarwalOpenXAITransparentEvaluation2024}.
Additionally, we provide the specificity, sensitivity, and class accuracy metrics to further examine the influences of the perturbation.


\subsubsection{PGI \& PGU}
First introduced in \cite{petsiukRISERandomizedInput2018} and later formalized in \cite{agarwalOpenXAITransparentEvaluation2024}, these metrics measure the difference between the original prediction and the prediction after an applied perturbation.
Based on the results from the explainability method, we determine the important and unimportant values, which are, in our case, the body key points with the highest and lowest SHAP values for the respective class. 
By perturbing the top-K important body key points for PGI and $\neg$ top-K body key points for PGU, the prediction gap between the unperturbed prediction $\hat{y} = f(x)$ and the perturbed prediction $f(x')$ can then measure the prediction difference.

\begin{equation}
    \text{PGI}(x, f, \mathbf{e}_x, k) = \mathbb{E}_{x' \sim \text{perturb}(x, \mathbf{e}_x, \text{top-}K)} \left[ | \hat{y} - f(x') | \right],
\end{equation}

\begin{equation}
\text{PGU}(x, f, \mathbf{e}_x, k) = \mathbb{E}_{x' \sim \text{perturb}(x, \mathbf{e}_x, \neg \text{top-}K)} \left[ | \hat{y} - f(x') | \right].
\end{equation}

Higher PGI values imply that perturbing important body key points results in performance degradation, implying that these are important for the model prediction.
PGU, on the contrary, captures the loss in performance when unimportant body key points are perturbed. 
Lower values indicate that the perturbed body key points are less influential in determining the outcome, thereby verifying the unimportance of these body key points. 
In summary, faithful importance attribution scores result in high values of PGI and low values of PGU.

\section{Experiments}
In this section, we conduct experiments on two real-world datasets to show the effectiveness of our proposed perturbation procedure and provide the SHAP explanations.
The arrangement of the experiment section is as follows.
First, we concisely present the used datasets and the implementation details.
Afterward, we will explain the two datasets and the importance of the features obtained with SHAP.
Furthermore, we describe the behavior of the metrics when informed and random perturbation is applied.

\subsection{Datasets}
\label{sec:dataset}
The NTU RGB+D 60 is a 3D human activity dataset for action recognition \cite{shahroudyNTURGB+DLarge2016}.
It consists of $56,880$ videos, holding 60 action classes. 
Our experiments use the defined subset cross-view (X-View) from \cite{shahroudyNTURGB+DLarge2016}.
This subset partitions the dataset according to camera perspectives. 
Here, camera views two and three are used as the training data $\mathcal{D}_{train}$, while camera view one is dedicated to the validation data $\mathcal{D}_{val}$.
The data is preprocessed by parallelizing the shoulder and hip joints, shifting the skeleton so that the spine joint is centered at the origin, scaling the lengths of the limbs, and normalizing the reference viewpoint.
Missing skeletons in the dataset are listed and provided by the authors from \cite{shahroudyNTURGB+DLarge2016}.
For our experiments, only actions displaying one person are enclosed.
The reader is referred to \cite{tempelAutoGCNgenericHumanActivity2024} for more details and the preprocessing steps for this dataset.

The second dataset comprises 557 videos of infants with medical risk factors for CP, recorded between 2001 and 2018 across four countries: the United States \((n=248)\), Norway \((n=190)\), Belgium \((n=37)\), and India \((n=82)\).
These recordings, including infant spontaneous movements in a supine position at 9 to 18 weeks, are standardized according to \cite{prechtlEarlyMarkerNeurological1997, einspielerPrechtlsMethodQualitative2004} and assessed by two medical experts.
For each frame within a video, the positions of 29 body key points of the infant in motion are captured as $X$ and $Y$ coordinates, resulting in the infant skeleton.
The video sequences are processed to improve their quality and consistency.
This involves resampling to 30 Hz, removing noise and inconsistencies in the data using a median filter, and adjusting the data to a common scale based on the trunk length as the reference point. 
The normalization is done by doubling the infant's trunk length and centering on the median mid-pelvis position. 
Finally, the processed data is divided into short windows, each lasting 5 seconds with a 2.5-second overlap.
The reader is referred to \cite{groosDevelopmentValidationDeep2022} for more details on the preprocessing and the dataset itself.

\begin{figure}[ht]
    \centering
    \begin{subfigure}[b]{0.3\textwidth}
        \centering
        \includegraphics[width=\textwidth]{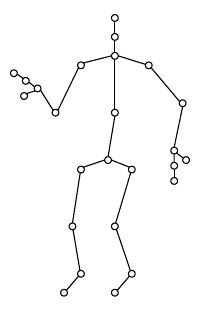} 
        \caption{NTU RGB+D skeleton}
        \label{fig:sub1}
    \end{subfigure}
    \hfill
    \begin{subfigure}[b]{0.3\textwidth}
        \centering
        \includegraphics[width=\textwidth]{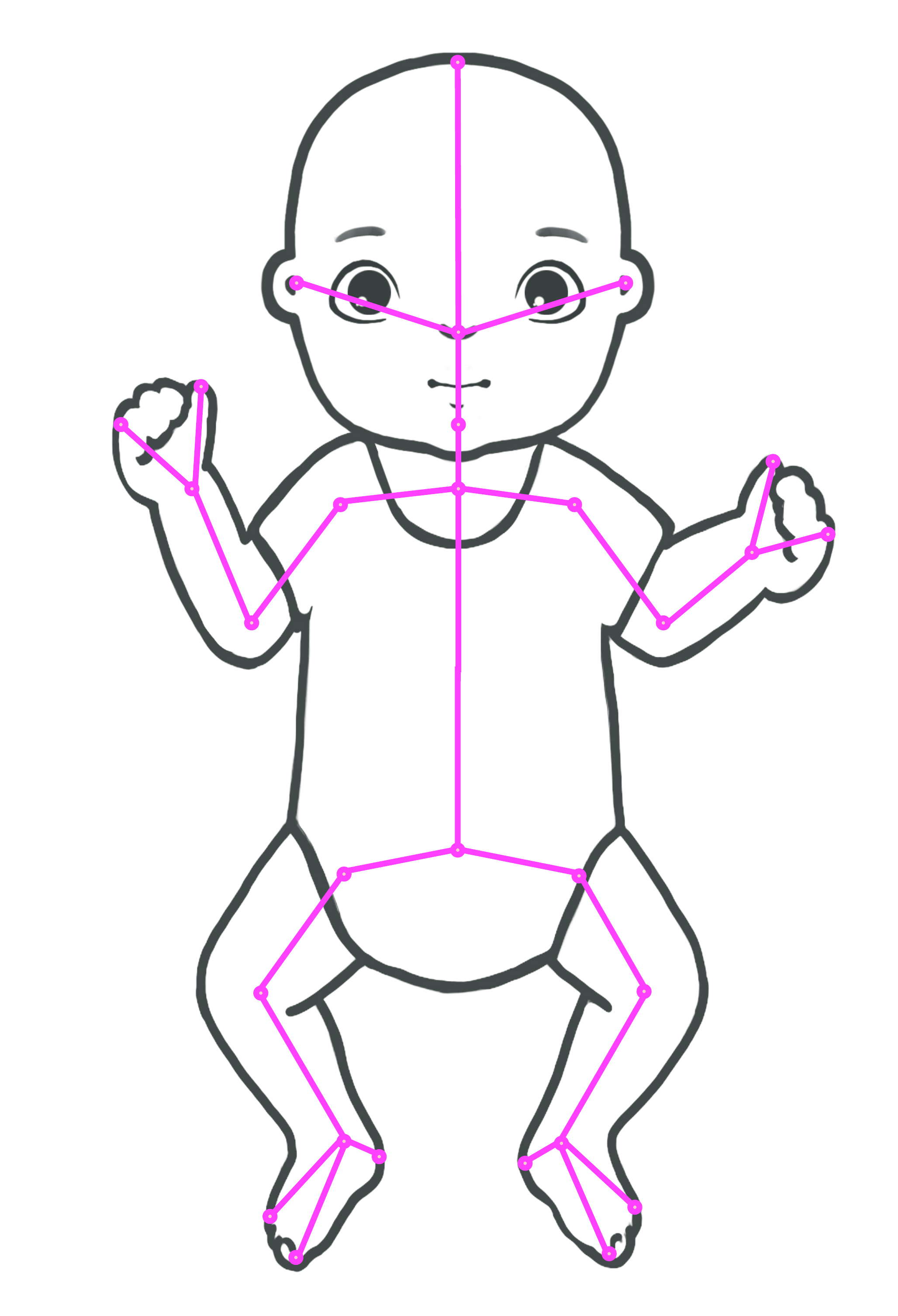} 
        \caption{CP skeleton}
        \label{fig:sub2}
    \end{subfigure}
    \hfill
    \begin{subfigure}[b]{0.3\textwidth}
        \centering
        \includegraphics[width=\textwidth]{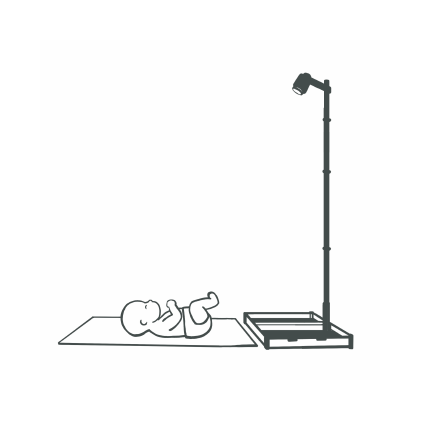} 
        \caption{CP recording environment}
        \label{fig:sub3}
    \end{subfigure}

    \caption{\textbf{Skeletons of the NTU RGB+D and CP dataset and the recording environment.} 
    (\subref{fig:sub1}) The NTU RGB+D skeleton with $n=19$ body key points. 
    (\subref{fig:sub2}) The CP skeleton with $n=29$ body key points adds finer detail in the toes and the head region.
    (\subref{fig:sub3}) The recording environment for the CP dataset.}
    \label{fig:skeleton}
\end{figure}

Each dataset requires a different dimensionality for $\mathbf{E}$ based on the number of body key points, ($n = 29$) for the CP dataset and ($n = 25$) for the NTU RGB+D dataset. 
Two example skeletons illustrating the edge connections for both datasets are presented in Fig.~\ref{fig:skeleton}, alongside the depiction of the recording environment used for the CP dataset.

\subsection{Implementation}
We use the best-performing model architecture for the NTU RGB+D 60 X-View dataset experiments from \cite{tempelAutoGCNgenericHumanActivity2024}.
Furthermore, we use the best-performing architecture from \cite{tempelLightweightNeuralArchitecture2024} for the CP dataset.
For the explanation, the DeepExplainer implementation from the SHAP library \cite{lundbergUnifiedApproachInterpreting2017} is adopted to calculate the SHAP values.
We define a random reference dataset with 100 instances possibly chosen from every class for the NTU RGB+D 60 dataset.
For the CP dataset, we generate a reference dataset consisting of 100 randomly chosen window samples from both classes.
As the GPU can only be used with a limited number of background samples, the experiments are conducted using subsamples ($n=20$) of the background, with the resulting SHAP values later aggregated.
The pseudocode for the implementation is presented in Algorithm \ref{alg:xai_algo}, where the input $\phi_0$ is calculated by SHAP and represents the average model prediction across the provided reference dataset, see Eq.~\ref{eq:phi_shap}.
The experiments are performed on a single NVIDIA-V100 with 32 GB GPU RAM on the PyTorch framework (version 2.3.1) \cite{paszkePyTorchImperativeStyle2019} with the global seed set to $1234$.
In our setup, computing SHAP values for the CP dataset (two classes) takes 16 hours and 15 minutes. 
For the NTU RGB+D dataset (three classes), the SHAP value calculation requires 23 hours and 41 minutes.

\begin{algorithm}
    \DontPrintSemicolon
    \caption{ShapGCN} \label{alg:xai_algo}
        \KwIn{Data: $\mathcal{D}_{\text{train}}$, $\mathcal{D}_{\text{val}}$; Model; $\phi_0$}
        \KwOut{$\phi$, $\mathcal{M}$}
        $\mathcal{M}    \gets \{\text{PGI, PGU}, \text{Specifitiy}, \text{Sensitvity}, \text{Accuracy} \}$\;
        \;
        Initialize trained model\;
        $\phi_0 = \mathbb{E}\left[\text{Model}(\mathcal{D_{\text{train}}}) \right]$\;
        \If{$\mathcal{D}$ is NTU RGB+D}{
            $p \gets \epsilon$
        }
        \Else{
            $p \gets 1 \times 10^{-5}$
        }
        $\phi \gets \textit{SHAP}(\mathcal{D}_{\text{val}}, \phi_0)$ \;

        \For{$k \gets 1$ \textbf{to} 10}{
            \tcp{Generate $\mathbf{E}$ for top-k most/least influential body key points}
            $\mathbf{E}_{\max \phi, k} \gets p$\;
            $\mathbf{E}_{\min \phi, k} \gets p$\;

            \For {d in $\mathcal{D}_{\text{val}}$}{
                $\mathcal{M}_{\max} \gets d, \mathbf{E}_{\max \phi, k}$\;
                $\mathcal{M}_{\min} \gets d, \mathbf{E}_{\min \phi, k}$\;
            }   
         }

\end{algorithm}

\subsection{Averaging Technique}
Given the multidimensional SHAP output, which matches the model input's dimensions, e.g., NTU RGB+D: $\mathbb{R}^{4 \times 6 \times 300 \times 25 \times 2 \times 60}$, CP: $\mathbb{R}^{4 \times 4 \times 225 \times 29 \times 1 \times 2}$, the SHAP values are averaged to generate interpretable explanations. The indices represent the following values:
\[[\text{Input feature} (f), \text{Feature dimension} (k), \text{Time frames} (t), \text{body key points} (n), \text{\# persons} (m), \text{SHAP values} (s)]\]

For the NTU RGB+D subset X-View, we compute the SHAP values for three randomly chosen classes and take the average over each.
To address padding, where classes repeat the same scene after the initial action sequence to maintain consistent input sizes, we reduce the temporal dimension ($t$) by half.
The input features ($f$) and feature dimensions ($k$) are merged into a single index. 
Additionally, we remove the second skeleton ($m$), focusing solely on the first skeleton by squeezing the dimensionality.
Thus, the new dimensionality for one skeleton sequence's SHAP values is $\mathbb{R}^{24 \times 150 \times 25 \times 60}$.

In the CP dataset, unlike NTU RGB+D, there is no second skeleton, so the index ($m$) is directly squeezed. 
As for the NTU RGB+D dataset, the input features ($f$) and feature dimensions ($k$) are combined into a single index. 
The resulting dimension for one window is $\mathbb{R}^{16 \times 225 \times 29 \times 2}$.

\subsection{CP dataset}

\subsubsection{Global Feature Importance}
\begin{figure}
    \centering
    \includegraphics[width=\textwidth]{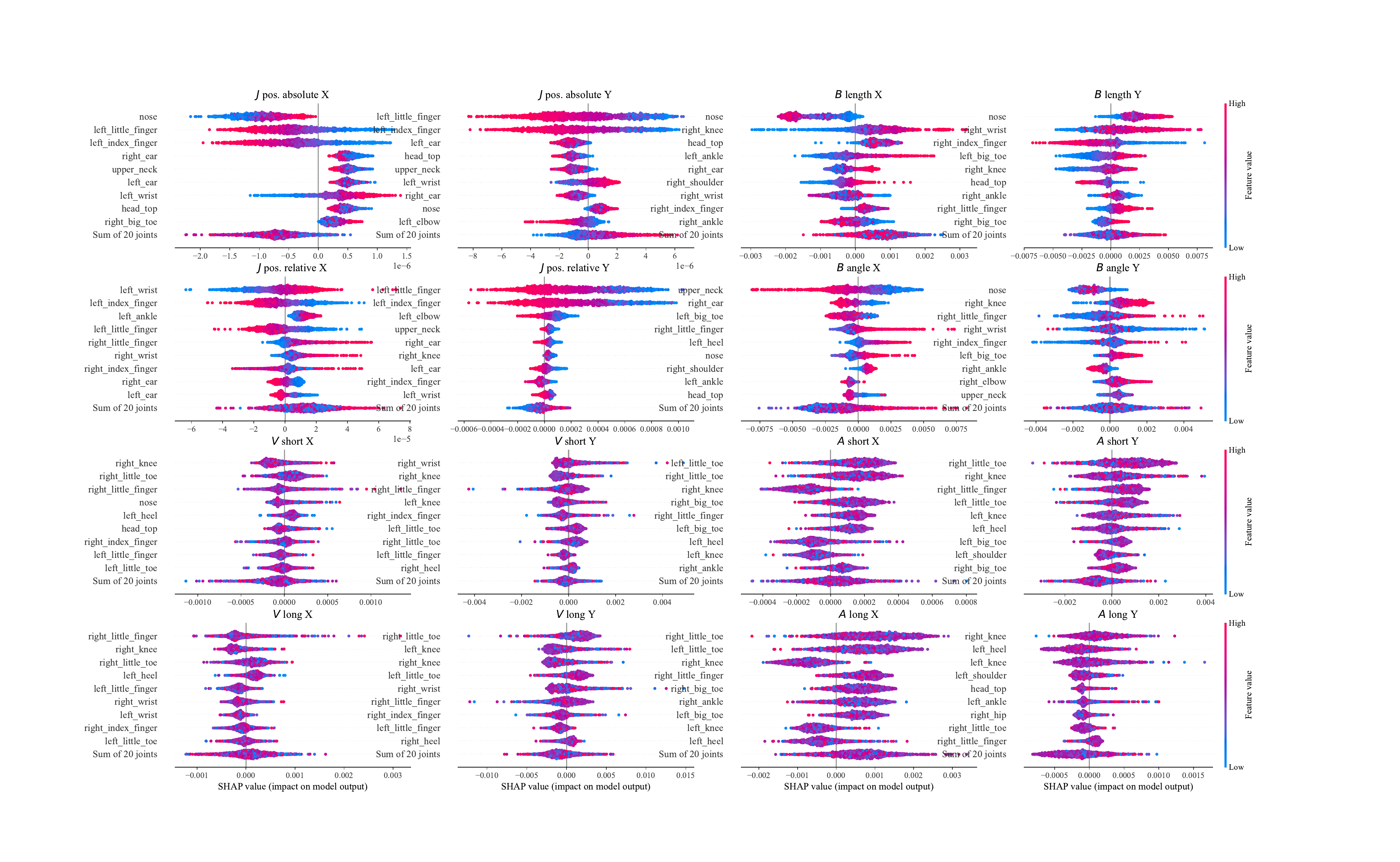}
    \caption{\textbf{SHAP values of the primary input features}.
    The figures show the SHAP value on the $x$-axis and the corresponding body key point on the $y$-axis, sorted in descending order of importance.
    The color of the dots indicates the feature value, with red corresponding to a high value, while blue is linked to a low feature value.
    The values are mean aggregated over the individual window time frames ($t$) and belong to class 1 (CP).}
    \label{fig:shaps_cp}
\end{figure}

To illustrate the global feature importance associated with CP predictions, we use the SHAP \textit{beeswarm} visualization \cite{lundbergUnifiedApproachInterpreting2017}. 
With this representation, the contributions of the specific input feature joint, velocity, bone, and acceleration $(J, V, B, A)$ for the validation dataset \(\mathcal{D}_{val}\) are highlighted. 
For this analysis, SHAP values are aggregated and averaged over the duration of each individual window. 
The resulting plot is presented in Fig.~\ref{fig:shaps_cp}.

The y-axis of each subplot lists the nine most important body key points for the respective feature in descending order of importance.
The importance is measured based on the average magnitude of the SHAP values across all windows. 
The 10th element represents a cumulative score that combines the SHAP values of all remaining body key points, capturing their collective contribution to the model output. 
The x-axis corresponds to the SHAP value, which reflects the influence of each feature on the model's output.
Positive values indicate a push toward the CP class (class 1), and negative values indicate a push toward the No-CP class (class 0).
The colored dots represent the actual values of the input features for each joint, where the color gradient provides a visual cue about the magnitude of the feature values, ranging from low to high across the body key points.

For features $J$ and $B$, a strong and consistent correlation is observed between the actual input feature values and their corresponding SHAP values across all windows. 
This implies that these features have a stable relationship with the model's prediction, where higher or lower values of $J$ and $B$ align with the likelihood of a particular class (CP or No-CP). 
This stability suggests that these features have a relatively straightforward and interpretable influence on the model's output across different windows from infants in the same class.

In contrast, the features $A$ and $V$ exhibit a more complex relationship. 
The SHAP values for these features do not consistently align with the input feature values across all windows. 
This lack of a clear, direct correlation is due to how these features are computed.
Acceleration and velocity have both positive and negative values, which introduces more variability in their contribution to the predictions. 
The distribution of the input feature values across the SHAP values for $A$ and $V$ suggests their influence on the model’s decision is more context-dependent.
Each individual window varies depending on the specific kinematics at this point. 
Because of this variability, the features $A$ and $V$ do not lend themselves to a simple, global interpretation like $J$ and $B$.
The contribution of $A$ and $V$ to the model's predictions may shift depending on the local patterns in the data, and a detailed, window-by-window exploration is necessary to fully understand how they impact the final predictions. 
Thus, while $J$ and $B$ provide consistent and interpretable information, $A$ and $V$ demand a more granular investigation to discern their roles in the model's decision-making process.

\subsubsection{Local Feature Importance}
The SHAP values for one randomly chosen infant with CP are aggregated and directly displayed on the respective body key points in the skeleton data to visualize the local explanation.

\begin{figure}
    \centering
    \includegraphics[width=\textwidth]{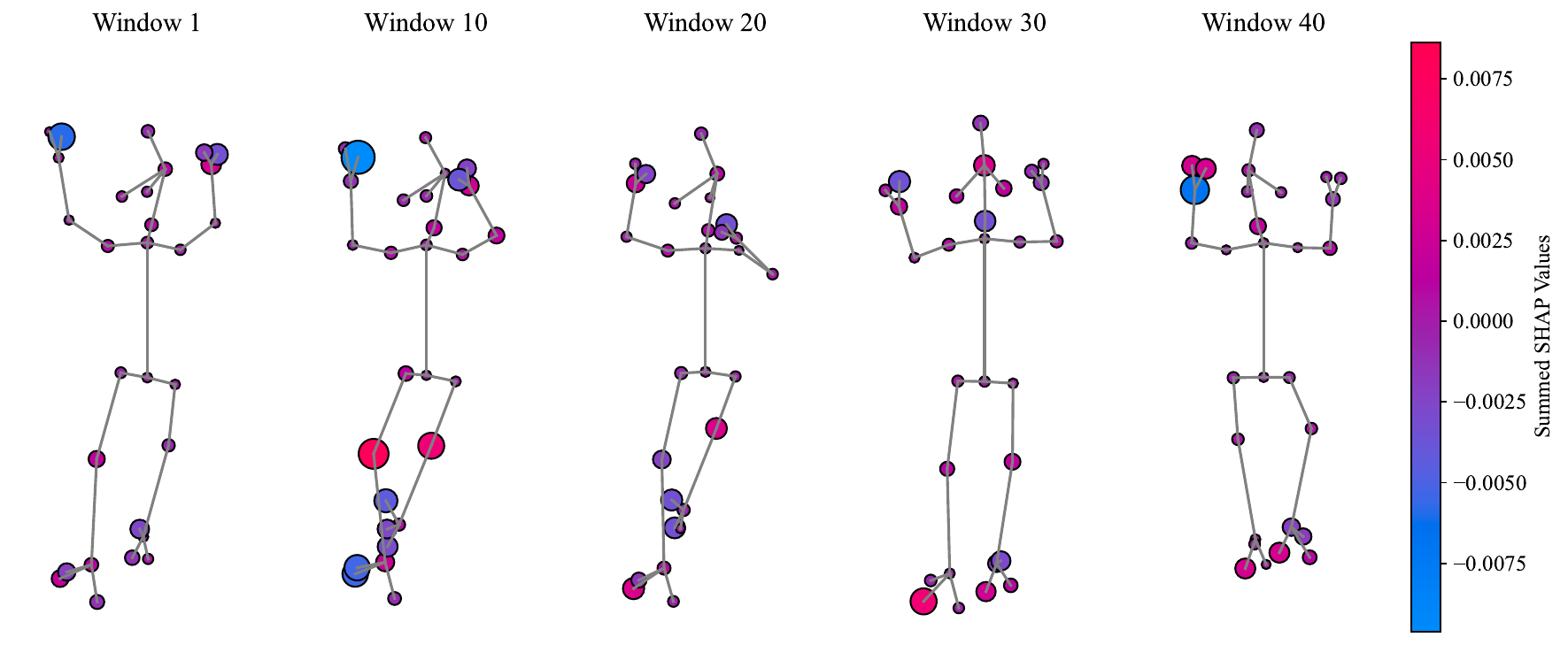}
    \caption{\textbf{Local SHAP values of the combined input features.} 
    The features ($J, V, B,$ and $A$) are added together and shown for five different windows of one infant with CP.
    The size and color of the dots indicate the value of the summed SHAP values.
    SHAP assigns the importance to the knees in window 10 and to the toes within windows 20, 30, and 40.
    Whereas a negative importance to CP is assigned to the hand region in windows 1 and 10.}
    \label{fig:shap_window_cp}
\end{figure}

Fig.~\ref{fig:shap_window_cp} illustrates the skeletal representation of the infant in five windows. 
In this visualization, the aggregated SHAP values from all features are mapped to the body key points at each time window. 
The size of each dot corresponds to the magnitude of the SHAP values, while the color indicates the direction of influence exerted by each joint. 
Specifically, red signifies a pull towards a prediction of CP, whereas blue indicates a pull towards No-CP. 
Body key points with SHAP values near zero are represented as smaller dots, highlighting their minimal influence on the prediction at the respective window.
The SHAP values indicate that the model output is most influenced by the infant’s four limbs freely moving against gravity in the supine position. 
The observation of smaller SHAP values in the head, trunk, and pelvis region is plausible because the infant is lying on its back during the video recording, resulting in minimal spatial movement from these body regions.

The local explanation of the individual features ($J, V, B,$ and $A$) for the same infant are provided in \ref{app:feature_cp}.

\subsubsection{Perturbation}
To assess the validity of the SHAP explanation, we perform a series of experiments with our proposed edge matrix perturbation.
In these experiments, the features derived from the dimension $(k)$, namely $x$ and $y$ orientations, are combined since these features are directly coupled. 
The mean of each body joint is calculated over the $(k)$ time frames of each window.
The obtained mean values are then aggregated on a joint-by-joint basis. 
Finally, the resulting values are sorted in descending order according to their corresponding SHAP values for each class.
Body key points identified as influential and non-influential for CP prediction are selected for modification, and these body key points are perturbed within the edge matrix $\mathbf{E}$ at their respective position $e_n$.

Table \ref{tab:metrics_cp_side_by_side} shows the perturbation results for the $\mathcal{D}_{val}$ CP dataset.
The table summarizes the performance metrics, comparing the effects of perturbing the top-K body key points identified as important to the two possible classification outcomes (CP vs.~No-CP).
This table provides insights into how an informed perturbation - particularly focusing on perturbing the top-K most influential body key points versus the least influential body key points - affects model performance regarding sensitivity, specificity, PGI, and PGU.
The first row is the baseline performance in which none of the body key points are perturbed, where the model achieves a sensitivity of $0.944$ and a specificity of $0.873$.
The two major columns describe the perturbation on the top-K body key points, either important or unimportant, for classifying CP or No-CP.
The arrows indicate the importance of those body key points towards one or the other class.

When focusing on the values in the first major column, in which the SHAP values have a low impact on CP and a high impact on No-CP, the $\text{PGI}_{\text{spec.}}$ remains relatively stable.
Even when more influential body key points are perturbed, the model maintains its ability to correctly identify No-CP cases, reflecting resilience in how the No-CP class is handled. 
This implies that the features critical for No-CP classification are either more evenly distributed across multiple body key points or less sensitive to the perturbations of the top-ranked body key points, allowing the model to recover from modifications to important body key points.

In the second major column, it can be observed that $\text{PGI}_{\text{sens.}}$ generally increases with the number of body key points perturbed, peaking at ten perturbed body key points ($\text{PGI}_{\text{sens.}} = 0.572$).
This indicates that perturbing the ten most important body key points leads to the most significant drop in performance, supporting the claim that these are crucial for correctly identifying CP.
Overall, the specificity for the most important body key points is fairly stable, ranging from 0.96 to 0.84, but the sensitivity decreases with each added joint, from $0.638$ to $0.372$. 
This confirms that the SHAP values correctly identify the body key points essential for classifying CP, as the model's performance on sensitivity is directly impacted when these body key points are perturbed.
Furthermore, the stability in specificity, despite the sensitivity drop, indicates that the model’s ability to classify non-CP cases correctly relies less on the same set of body key points.

\begin{table}[]
    \centering
    \caption{\textbf{Performance metrics on the CP dataset under various perturbation scenarios.}
    The table compares the effects of perturbing the most important body key points versus the least important ones, as determined by the respective SHAP value for this specific window. 
    Different numbers of top-K influential body key points are perturbed, with the perturbation value for the edge matrix set to $1e-5$, simulating a masking effect. 
    Due to the class imbalance in the dataset - where healthy infants outnumber those with CP - both specificity and sensitivity metrics are reported for each perturbation scenario. 
    The $\text{PGI}$ and $\text{PGU}$ values are reported based on important and unimportant body key points for the respective setting gains/losses under perturbation - higher values for PGI imply greater explanation faithfulness while it is the opposite for PGU.}
    \label{tab:metrics_cp_side_by_side}
    \resizebox{\linewidth}{!}{%
    \begin{tabular}{@{}lllllllll@{}}
        \toprule
        \# top-K  & \multicolumn{4}{c}{SHAP value CP $(\downarrow)$ No-CP $(\uparrow)$} & \multicolumn{4}{c}{SHAP value CP $(\uparrow)$ No-CP $(\downarrow)$} \\
        \cmidrule(lr){2-5} \cmidrule(lr){6-9}
        body key points & $\text{PGI}_{\text{spec.}} (\uparrow)$ & $\text{PGU}_{\text{sens.}} (\downarrow)$& Specificity & Sensitivity & $\text{PGI}_{\text{sens.}} (\uparrow)$ & $\text{PGU}_{\text{spec.}} (\downarrow)$ & Specificity & Sensitivity \\ 
        \midrule
        No perturb. & - & - & 0.873 & 0.944 & - & - & 0.873 & 0.944 \\

        1 & 0.108 & 0.234 & 0.765 & 0.710 & 0.306 & 0.080 & 0.793 & 0.638\\
        2 & 0.137 & \textbf{0.215} & 0.736 & 0.729 & 0.326 & 0.094 & 0.779 & 0.618\\
        3 & 0.152 & 0.223 & 0.721 & 0.721 & 0.362 & 0.104 & 0.769 & 0.582\\
        4 & \textbf{0.157} & 0.239 & 0.716 & 0.705 & 0.395 & 0.113 & 0.760 & 0.549\\
        5 & 0.154 & 0.255 & 0.719 & 0.689 & 0.429 & 0.108 & 0.765 & 0.515\\
        6 & 0.143 & 0.257 & 0.730 & 0.687 & 0.440 & 0.100 & 0.773 & 0.504\\
        7 & 0.137 & 0.263 & 0.736 & 0.681 & 0.477 & 0.099 & 0.774 & 0.467\\
        8 & 0.131 & 0.276 & 0.742 & 0.668 & 0.505 & 0.084 & 0.789 & 0.439\\
        9 & 0.126 & 0.298 & 0.747 & 0.646 & 0.532 & 0.069 & 0.804 & 0.412\\
        10 & 0.120 & 0.309 & 0.753 & 0.635 & \textbf{0.572} & \textbf{0.060} & 0.813 & 0.372\\

        \bottomrule
    \end{tabular}
    }
\end{table}

\subsubsection{Random Perturbation}
Random perturbation is a control to evaluate the perturbations in the informed setting.
This method randomly perturbs body key points without relying on SHAP values. 
The goal is to determine whether targeting specific body key points based on their importance (as in the SHAP-guided, informed perturbation) leads to significantly different outcomes than randomly perturbing body key points.

From Table \ref{tab:metrics_random}, we observe that random perturbations have a milder effect on both specificity and sensitivity when compared to informed perturbations from Table \ref{tab:metrics_cp_side_by_side}. 
For instance, the sensitivity under random perturbation decreases to $0.612$ at most, whereas in Table \ref{tab:metrics_cp_side_by_side}, informed perturbation for the CP class reduces the sensitivity to $0.372$. 
This indicates that perturbing the most important body key points, identified through the SHAP values, has a far more detrimental impact on the model's performance. 
Specifically, informed perturbation leads to more significant drops in sensitivity, particularly when perturbing the top 10 most important body key points, indicating that these body key points hold critical information for an accurate prediction of CP.

This supports the validity of using SHAP-based explanations, as perturbing the most important body key points meaningfully degrades the model’s predictive performance. 
Hence, informed perturbation provides a more faithful assessment of the model’s reliance on the specific features at a specific body key point.

\begin{table}[]
    \centering
    \caption{\textbf{Performance metrics on the CP dataset with random perturbation.} 
    The table displays the effects of randomly perturbing a specified number of body key points on specificity and sensitivity.}
    \label{tab:metrics_random}
    \begin{tabular}{@{}lll@{}}
        \toprule
        \# body key points & \multicolumn{2}{c}{Random Perturbation}  \\
                        \cmidrule(lr){2-3}  
                  & Specificity & Sensitivity \\ 
        \midrule
        No perturb. & 0.873 & 0.944 \\
        1  & 0.792 & 0.728 \\
        2  & 0.789 & 0.711 \\
        3  & 0.790 & 0.708 \\
        4  & 0.789 & 0.691 \\
        5  & 0.789 & 0.696 \\
        6  & 0.790 & 0.680 \\
        7  & 0.793 & 0.664 \\
        8  & 0.790 & 0.641 \\
        9  & 0.800 & 0.648 \\
        10 & 0.804 & 0.612 \\
        \bottomrule
    \end{tabular}
\end{table}

\subsection{NTU RGB+D}
\subsubsection{Global Feature Importances}
\begin{figure}
    \centering
    \includegraphics[width=\textwidth]{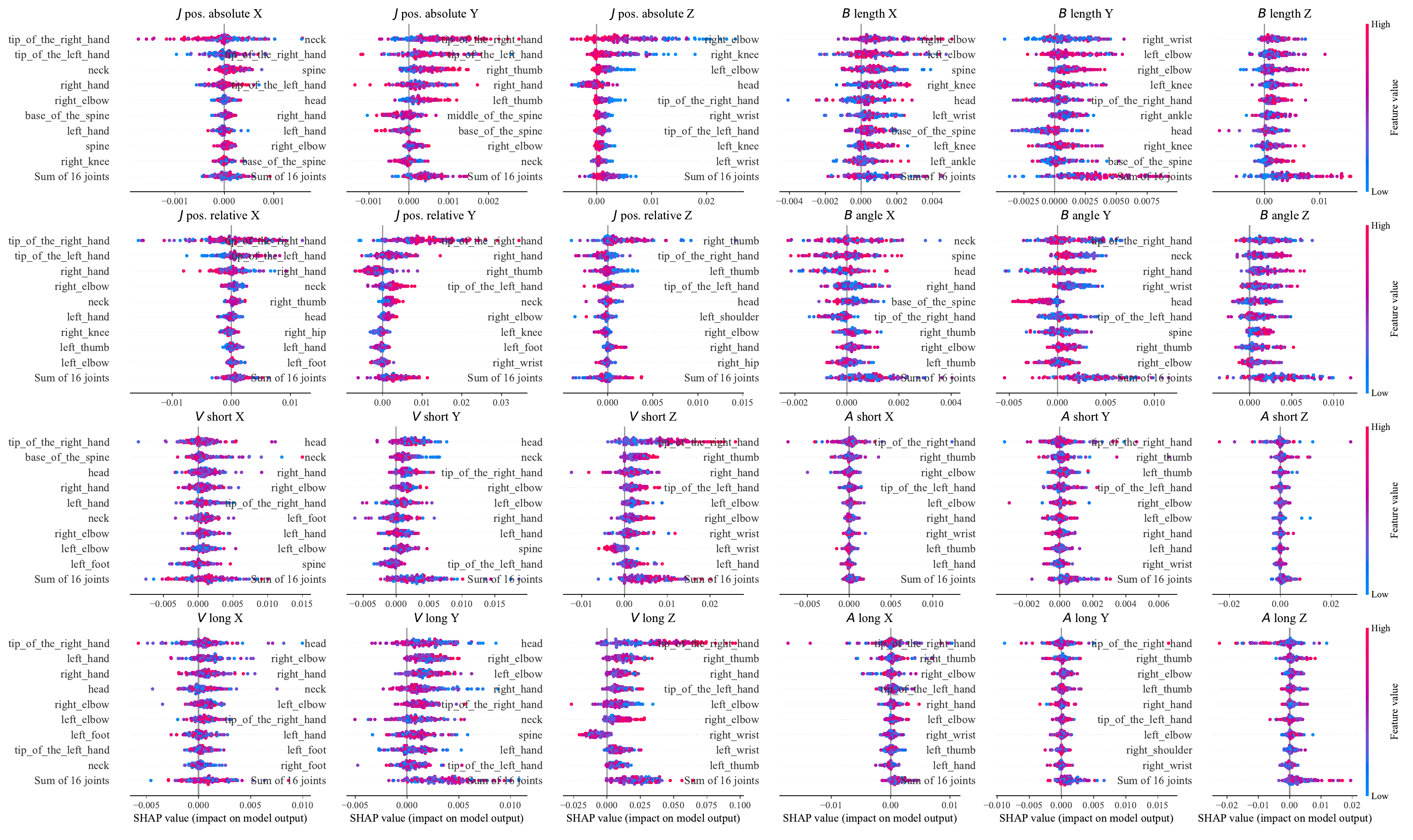}
    \caption{\textbf{SHAP values of the primary input features}.
    The figures show the SHAP value on the $x$-axis and the corresponding body key point on the $y$-axis, sorted in descending order of importance.
    The color of the dots indicates the feature value, with red corresponding to a high value, while blue is linked to a low feature value.
    The values are mean aggregated over the individuals performing the action and belong to class 6 (\textit{pick up}).}
    \label{fig:shaps_ntu}
\end{figure}

We show the global feature importance for class 6 in the NTU RGB+D X-View subset, corresponding to the action \textit{pick up}.
As for the CP dataset, we utilize the SHAP \textit{beeswarm} plot to show the feature pulls across the body key points for the individual input features.
In Fig.~\ref{fig:shaps_ntu}, the SHAP values on the individual input features are shown, where we have the additional coordinate orientation $Z$.

For the input features $J$ and $B$, the features show a separation and noticeable trend across the body key points.
This trend, however, is not as distinctive as for the CP dataset, where the windows show a clear correlation between the SHAP and feature values.

Furthermore, it can be observed that the feature and SHAP values are not as distinctly correlated for the input features $V$ and $A$. 
For example, the SHAP values for the $A$ \textit{short} along the $X$, $Y$, and $Z$ axes show a distribution where the contribution of individual key points is spread evenly across the shown body key points, and no dominant body key point clearly stands out.
Similarly, for the input features $V$, different axes exhibit variability in the SHAP and feature values. 
The SHAP values do not correlate strongly and consistently with the actual input values across the individual time points.
This highlights the more complex nature of how $A$ and $V$ contribute to the model’s predictions for this dataset, suggesting that these features likely capture more nuanced kinematics that are harder to interpret globally without considering the specific context and time point where the action occurs.

This is a notable difference from the CP dataset, where specific key points exhibited a more pronounced relationship with the SHAP values of these features, and is due to the long time frame ($t$) on which the SHAP values are computed for the NTU RGB+D dataset.
However, it can be observed that the body joints of the right and left hand appear to be the most important feature in almost all feature dimensions.
This is plausible since the action \textit{pick up} is mainly executed by those limbs.

\subsubsection{Local Feature Importance}
\begin{figure}
    \centering
    \includegraphics[width=\textwidth]{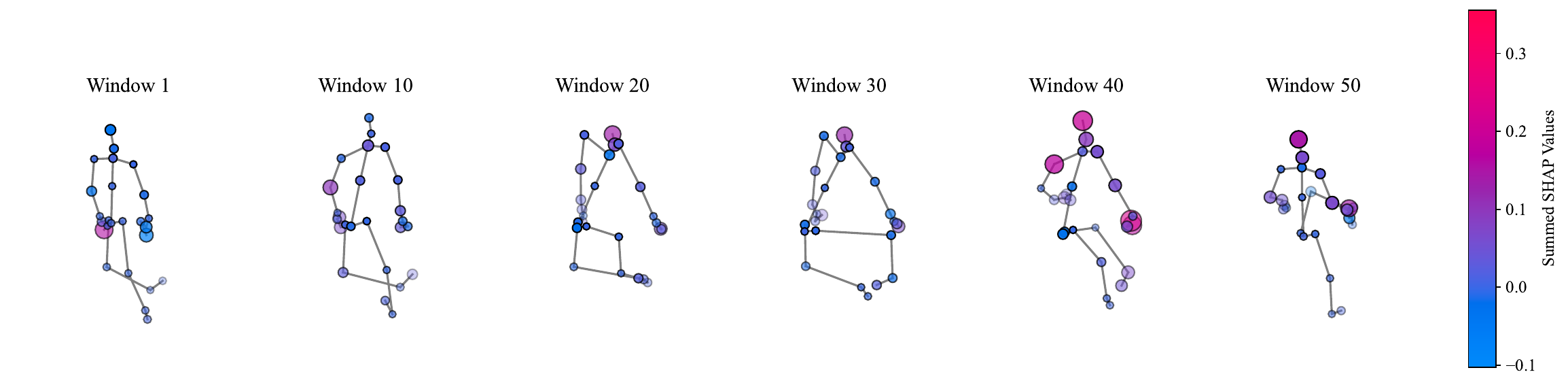}
    \caption{\textbf{Local SHAP values of the combined input features for class 6 (\textit{pick up}).} 
    The features ($J, V, B,$ and $A$) are added together and shown for five different windows.
    The size and color of the dots indicate the value of the summed SHAP values.
    SHAP assigns the highest importance for the action to the hand and head region in windows 40 and 50, while other body key points remain neutral (with SHAP values near zero) throughout the remaining windows.}
    \label{fig:shap_windows_ntu}
\end{figure}

The SHAP values for a randomly selected individual from class 6 (\textit{pick up}) are aggregated and visually represented on the corresponding body key points within the skeleton data.
This allows for a localized interpretation similar to that of the CP dataset. 
Figure~\ref{fig:shap_windows_ntu} shows this skeletal representation of the person performing the action across six Windows.

This visualization maps the aggregated SHAP values derived from all features onto the body key points for each time window. 
The size of each dot reflects the magnitude of the SHAP values, while the color indicates the direction of influence of each joint. 
Specifically, red dots signify a pull towards class 6, whereas blue dots indicate a pull away from this class.
Body key points associated with SHAP values close to zero are represented as smaller dots, thereby emphasizing their minimal impact on the prediction for the respective window. Notably, the SHAP values suggest that the model's output for this individual is predominantly influenced by the hands engaged in the action during window 40. 
The observed smaller SHAP values in Windows 1 and 10 are reasonable, as the action has not yet commenced during these intervals.

\subsubsection{Perturbation}
As with the CP dataset, we assess the validity of the obtained SHAP explanations by perturbing the influential and non-influential body key points.
Unlike the CP dataset, where action sequences are segmented into 5-second windows, the NTU RGB+D dataset stores each action sequence as a single continuous time frame.
Given the dataset's sensitivity to edge masking, where fixing edges to near-zero values greatly impacts the model prediction, we do the perturbations by adjusting edge weights with a percentage-based threshold.

Table \ref{tab:perturb_xview} reports the results for perturbing the edges with a $\epsilon = 35 \%$ change.
In \ref{app:ntu_thresh}, we show the perturbation experiment when the edges are perturbed with threshold changes from $\epsilon \in [0.05, 0.45]$.
This systematic manipulation of body key points allows us to investigate how perturbations influence the model prediction based on the importance of features as determined by SHAP.

Notably, when body key points associated with positive SHAP values are perturbed, a sharp decline in classification accuracy is observed across all the shown actions.
Conversely, although perturbing joints classified as non-influential also result in a decline in accuracy, this decline is less severe than that observed with influential joints.
This trend is plausible and suggests that while the accuracy decreases for unimportant joints, the impact is moderated relative to the notable effect of perturbing important joints.
Furthermore, across all three classes, the PGI value is constantly higher than the PGU value, which implies that the body key points determined by SHAP are truly important or unimportant.
These findings highlight the efficacy of SHAP in identifying critical body key points in this dataset.

\begin{table}
    \centering
    \caption{\textbf{Performance metrics on the NTU RGB+D dataset under informed perturbation scenarios.}
    The accuracy progressively decreases for each class as the top-K body points are perturbed. 
    We show the impact of perturbing important (imp.) vs. unimportant ($\neg$ imp.) body key points with the threshold set to $\epsilon=35\%$ on classification accuracy for three action classes and report the PGI and PGU metrics.
    Accuracy drops significantly in all three classes when perturbing influential points, confirming SHAP’s ability to correctly identify crucial body key points.}

    \label{tab:perturb_xview}
    \resizebox{\linewidth}{!}{%
    \begin{tabular}{@{}lllllllllllll@{}}
        \toprule
        
        \# top-K  & \multicolumn{4}{c}{Class 6 (\textit{pick up})} & \multicolumn{4}{c}{Class 11 (\textit{reading})} & \multicolumn{4}{c}{Class 16 (\textit{put on a shoe})} \\
        \cmidrule(lr){2-5} \cmidrule(lr){6-9} \cmidrule(lr){10-13}
        body key points & Acc. imp. & Acc. $\neg$ imp. & PGI $(\uparrow)$ & PGU $(\downarrow)$ & Acc. imp. & Acc. $\neg$ imp. & PGI $(\uparrow)$ & PGU $(\downarrow)$ & Acc. imp. & Acc. $\neg$ imp. & PGI $(\uparrow)$ & PGU $(\downarrow)$\\ 
        \midrule
    
        No perturb. & 0.971 & 0.971 & - & - & 0.797 & 0.797 & - & - & 0.845 & 0.845 & - & -  \\
        1 & 0.883 & 0.940 & 0.088 & \textbf{0.031} & 0.597 & 0.733 & 0.200 & \textbf{0.064} & 0.533 & 0.784 & 0.312 & \textbf{0.061} \\   
        2 & 0.851 & 0.918 & 0.120 & 0.053 & 0.568 & 0.667 & 0.229 & 0.130 & 0.444 & 0.727 & 0.401 & 0.118 \\
        3 & 0.759 & 0.883 & 0.212 & 0.088 & 0.505 & 0.632 & 0.292 & 0.165 & 0.406 & 0.686 & 0.439 & 0.159 \\  
        4 & 0.617 & 0.839 & 0.354 & 0.132 & 0.470 & 0.568 & 0.327 & 0.229 & 0.397 & 0.629 & \textbf{0.448} & 0.216 \\   
        5 & 0.560 & 0.807 & 0.411 & 0.164 & 0.371 & 0.486 & 0.426 & 0.311 & 0.400 & 0.575 & 0.445 & 0.270 \\
        6 & 0.541 & 0.782 & \textbf{0.430} & 0.189 & 0.305 & 0.397 & 0.492 & 0.400 & 0.429 & 0.540 & 0.416 & 0.305 \\ 
        7 & 0.566 & 0.772 & 0.405 & 0.199 & 0.232 & 0.368 & 0.565 & 0.429 & 0.406 & 0.521 & 0.439 & 0.324 \\  
        8 & 0.573 & 0.763 & 0.398 & 0.208 & 0.187 & 0.337 & 0.610 & 0.460 & 0.422 & 0.511 & 0.423 & 0.334 \\  
        9 & 0.557 & 0.753 & 0.414 & 0.218 & 0.117 & 0.321 & 0.680 & 0.476 & 0.432 & 0.505 & 0.413 & 0.340 \\ 
       10 & 0.544 & 0.734 & 0.427 & 0.237 & 0.086 & 0.314 & \textbf{0.711} & 0.483 & 0.416 & 0.502 & 0.429 & 0.343 \\

        \bottomrule
    \end{tabular}%
    }
\end{table}

\subsubsection{Random Perturbation}
We run the same experiment by randomly perturbing the edges to further validate our proposed perturbation approach.
Table \ref{tab:ntu_random} shows the results of perturbing between one to ten body key points for classes 6, 11, and 16 randomly.

From Table \ref{tab:ntu_random}, it is observable that random perturbations induce a more gradual decline in accuracy compared to informed perturbations from Table \ref{tab:perturb_xview}. 
This can be attributed to the fact that, in the random case, both influential and non-influential body key points have equal chances of being perturbed, leading to a diluted effect.
For instance, with the SHAP-guided perturbations, accuracy decreases sharply when critical body key points are disrupted, as seen in Class 6 (\textit{pick up}), where perturbing five influential points drops accuracy from $0.971$ to $0.560$. 
In contrast, random perturbations reduce accuracy more moderately, with only a $0.227$ drop in accuracy after perturbing five body key points. 

However, random perturbations eventually also lead to substantial accuracy degradation, as seen when perturbing a larger number of body key points (e.g., 10), where the model’s accuracy diminishes almost to the same extent as with the informed perturbation scenario. 
This behavior illustrates that while random perturbations are initially less disruptive, the model becomes increasingly unstable as more body key points are altered.

\begin{table}
    \centering
    \caption{\textbf{Performance metrics on the NTU RGB+D dataset with random perturbation.} 
    The effects of randomly perturbing a specified number of body key points on the class accuracy of classes 6 (\textit{pick up}), 11 (\textit{reading}), and 16 (\textit{put on a shoe}).}
    \label{tab:ntu_random}
    \begin{tabular}{@{}llll@{}}
        \toprule
        \# body key points & \multicolumn{3}{c}{Class Accuracy}  \\
            \cmidrule(lr){2-4}  
                  & Class 6 & Class 11 & Class 16 \\ 
        \midrule
        No perturb. & 0.971 & 0.797 & 0.845  \\
        1 & 0.956 & 0.778 & 0.790 \\  
        2 & 0.937 & 0.702 & 0.775 \\  
        3 & 0.886 & 0.616 & 0.724 \\  
        4 & 0.769 & 0.524 & 0.679 \\
        5 & 0.744 & 0.483 & 0.568 \\ 
        6 & 0.639 & 0.321 & 0.508 \\  
        7 & 0.541 & 0.283 & 0.489 \\  
        8 & 0.459 & 0.244 & 0.406 \\  
        9 & 0.361 & 0.159 & 0.356 \\  
        10 & 0.294 & 0.089 & 0.283 \\  
        \bottomrule
    \end{tabular}
\end{table}

\section{Discussion}
The results of this study constitute significant progress towards understanding how SHAP can be used to explain GCNs in HAR.
With the novel perturbation strategy, we can selectively alter the most important body key points identified through SHAP values.
We do this by probing the model's reliance on the features associated with the highest SHAP value in the two datasets, i.e., \ perturbing the most important body key points for the model. 
This leads to a sharp decline in sensitivity for the CP dataset, which measures the ability of the model to correctly identify positive cases among all actual positives. 
For the NTU RGB+D dataset, we see a decrease in class accuracy for the individual action.
This serves as a strong verification of SHAP’s role in offering faithful explanations, as the features it deemed most important are indeed those that, when perturbed, most affect the model's prediction outcome. 
As a baseline, we show that random perturbation has a significantly lesser effect on the model’s performance: When body key points are perturbed randomly rather than based on SHAP values, the model’s sensitivity and specificity decline more moderately. 
This contrast between random and SHAP-guided perturbations indicates that SHAP indeed effectively identifies important features.

Trust from the users in ML models is important, especially in medical settings \cite{bharatiReviewExplainableArtificial2024, chaddadSurveyExplainableAI2023, tjoaSurveyExplainableArtificial2021}. 
Our findings show that HAR models in clinical applications need not be considered ``black boxes''.
With ShapGCN, we lay the foundation for clinicians to understand why a model makes its specific decisions. 
This way, clinicians can gain trust in ML models by understanding how important different input features are for the model, retaining the ability to choose \emph{not} to trust the model, e.g., \ if it makes use of input features the clinician knows to be unimportant.
However, to ensure that the provided explanations are genuinely useful, it is essential to research what specific information clinicians need and how they prefer it to be visualized or communicated. 
Tailoring these insights to their requirements can significantly enhance their trust and make the tool more practical within the clinical routine.
For instance, if certain body key points or limb areas are consistently flagged as important for diagnosing CP, this could lead to more targeted therapies or insights into the diagnostic processes.

While our perturbation strategy works well within our GCN framework, it is not guaranteed to generalize across other architectures or datasets. 
Other perturbation techniques on the input data, such as a naive perturbation approach of zeroing the coordinate orientation of the body key points, would collapse the body key points at the origin. 
This renders the data representations meaningless and distorts the spatial relationships within the skeleton too much for an informed perturbation.
Further investigation is needed to explore whether matching perturbation strategies can be successfully applied to other data types, such as RGB video or time-series sensor data, as often used in HAR. 
Finally, additional datasets should be tested to validate our proposed method further.

The scope of perturbation techniques within SHAP is constrained to post-hoc methods.
This hinders its deployment in dynamic or real-time systems, where the immediate interpretation of prediction results is required.
A notable challenge associated with the use of SHAP is its high computational cost, particularly when processing large datasets. 
This computational burden originates from the computational complexity of calculating Shapley values, which presents a challenge to the scalability of the approach for real-time or large-scale applications. 
This issue can be resolved by parallelizing the SHAP calculation on different nodes, but it requires more computational resources, which may not be available in every setting.
Additionally, approximation techniques like ``Kernel-'', ``Tree-'', and ``Deep SHAP'' can be used to speed the process of obtaining the explanation \cite{lundbergUnifiedApproachInterpreting2017}.

SHAP explanations offer spatial-temporal interpretations of the impact of different skeleton body key points. 
However, the size and composition of the reference dataset have an impact on these explanations. 
Selecting an appropriate reference dataset that reflects the dataset distribution is crucial to ensure reliable and meaningful explanations.
Although a larger reference dataset offers a more detailed explanation, it also requires more computational effort, increasing exponentially.
In this context, loading the entire dataset combined with a large model on the GPU may be currently not possible when resources are restricted, necessitating several iterations to obtain the explanations.
GradCAM may be the better alternative in an environment with resource restrictions due to its lower computational requirements, as shown in \cite{tempelChooseYourExplanation2024}.

Biomarkers depict unique features or patterns tied to specific movements and hold critical importance in HAR.
This is especially important within clinical applications such as the CP dataset is used.
SHAP offers insights into the contribution of input features and has the potential to identify biomarkers.
Due to the cooperative game theory concept, SHAP's close relationship with the input data makes it a promising tool for biomarker identification. 
Identifying these biomarkers with SHAP could help diagnose and monitor infant conditions.
SHAP can also serve as a valuable method for discovering how certain input features contribute to recognizing such movement-related biomarkers.
However, this concept remains unexplored in this work and requires further validation within a clinical study.

Finally, the averaging technique for the SHAP values may obscure nuanced yet important variations in data.
This might be especially important within the windows of the CP dataset, where absent fidgety movements can indicate CP 
\cite{einspielerPrechtlsMethodQualitative2004}. 
For instance, subtle short temporal dynamics in the windows may be lost through averaging and potentially impact the interpretability of these essential patterns.
Future work could refine this technique to preserve such details and offer more exhaustive explanations.
However, which information is finally communicated to the clinicians has to be decided.

\section{Conclusion}
This study highlights the efficacy of SHAP in explaining the decision-making processes of GCNs used in HAR tasks based on skeleton data. 
Our novel perturbation strategy enables examining how specific body key points influence model predictions, and we identify the importance of body key points to the model on two real-world datasets, classifying CP and human actions. 
Perturbing the most important body key points significantly compromises the model's sensitivity on the CP dataset and class accuracy on the model for the NTU RGB+D dataset, validating the importance of the specific body key points as obtained using SHAP.
This reinforces the reliability of SHAP values as an attribution method for feature importance.
We can also gain granular insight into the specific feature contributions to the model output through SHAP.

While our findings are promising, they also have some limitations, including the computational burden associated with SHAP calculations and the potential lack of generalizability of the perturbation technique across diverse datasets and tasks. 
Adapting our perturbation approach to different datasets and architectures and exploring the use of SHAP in biomarker identification are promising areas for future research. 
Our study lays a foundation for further advances in the explainability of models used in HAR, especially in the medical domain.

\section*{Contributions}
\textbf{Felix Tempel:} Conceptualization; Methodology; Formal analysis; Software; Writing - original draft preparation.
\textbf{Espen Alexander F. Ihlen:} Data curation; Supervision; Writing - review and editing.
\textbf{Lars Adde:} Data curation; Writing - review and editing.
\textbf{Inga Strümke:} Project administration; Supervision; Writing - review and editing.

\section*{Data Availability}
The NTU RGB+D 60 dataset that supports the findings of this study is available in \url{https://rose1.ntu.edu.sg/dataset/actionRecognition/} \cite{shahroudyNTURGB+DLarge2016}.
Due to ethical considerations, the CP dataset is not publicly available.

\section*{Code availability}
The code and the experimental results are publicly available at \url{https://github.com/DeepInMotion/ShapGCN}.

\section*{Acknowledgements}
This work was partly supported by the PERSEUS Project, a European Union’s Horizon 2020 Research and Innovation Program under the Marie Skłodowska-Curie under Grant 101034240, and partly by the Research Council of Norway under Grant 327146.

\printbibliography

\appendix

\section{Individual Feature Importance CP dataset}
\label{app:feature_cp}

\begin{figure}[]
    \centering
    \begin{subfigure}[b]{0.67\textwidth}
        \centering
        \includegraphics[width=\textwidth]{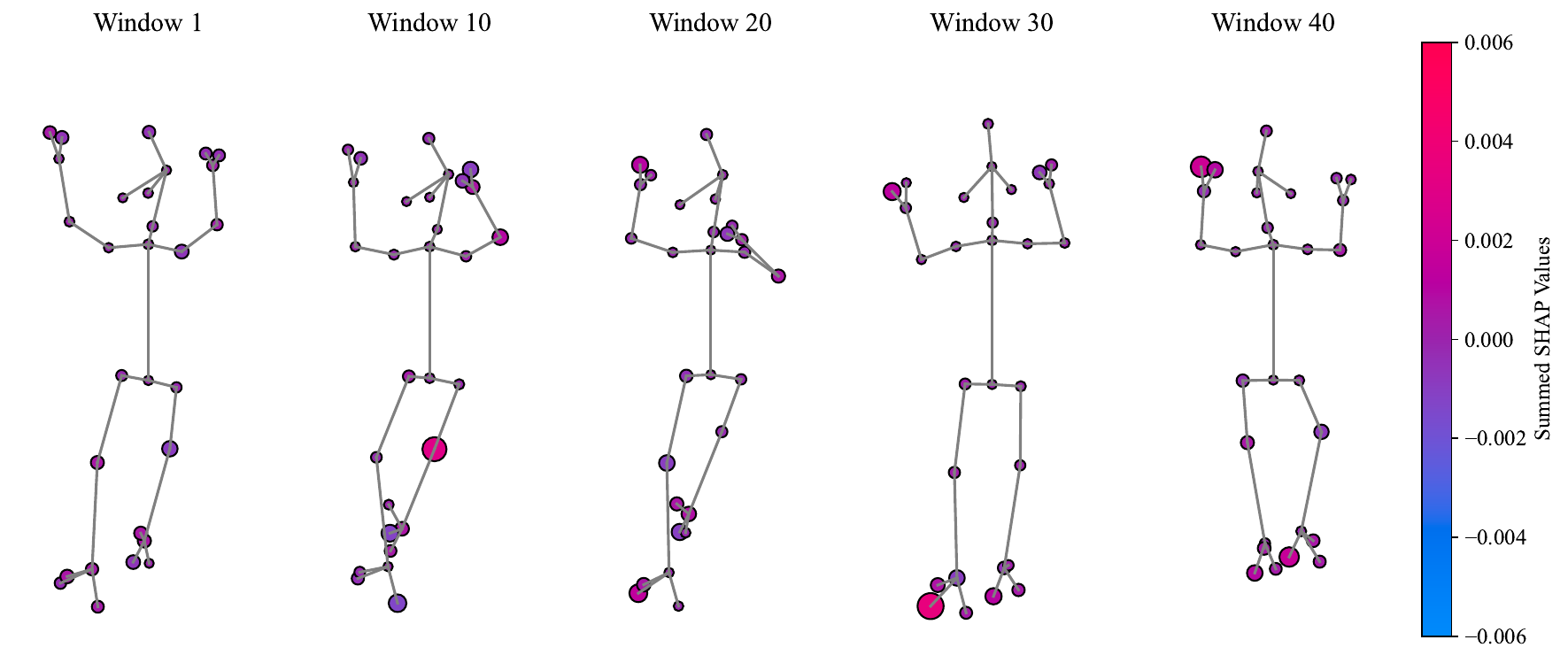}
        \caption{SHAP values of the acceleration feature.}
        \label{fig:shap_window_acc}
    \end{subfigure}
    
    \begin{subfigure}[b]{0.67\textwidth}
        \centering
        \includegraphics[width=\textwidth]{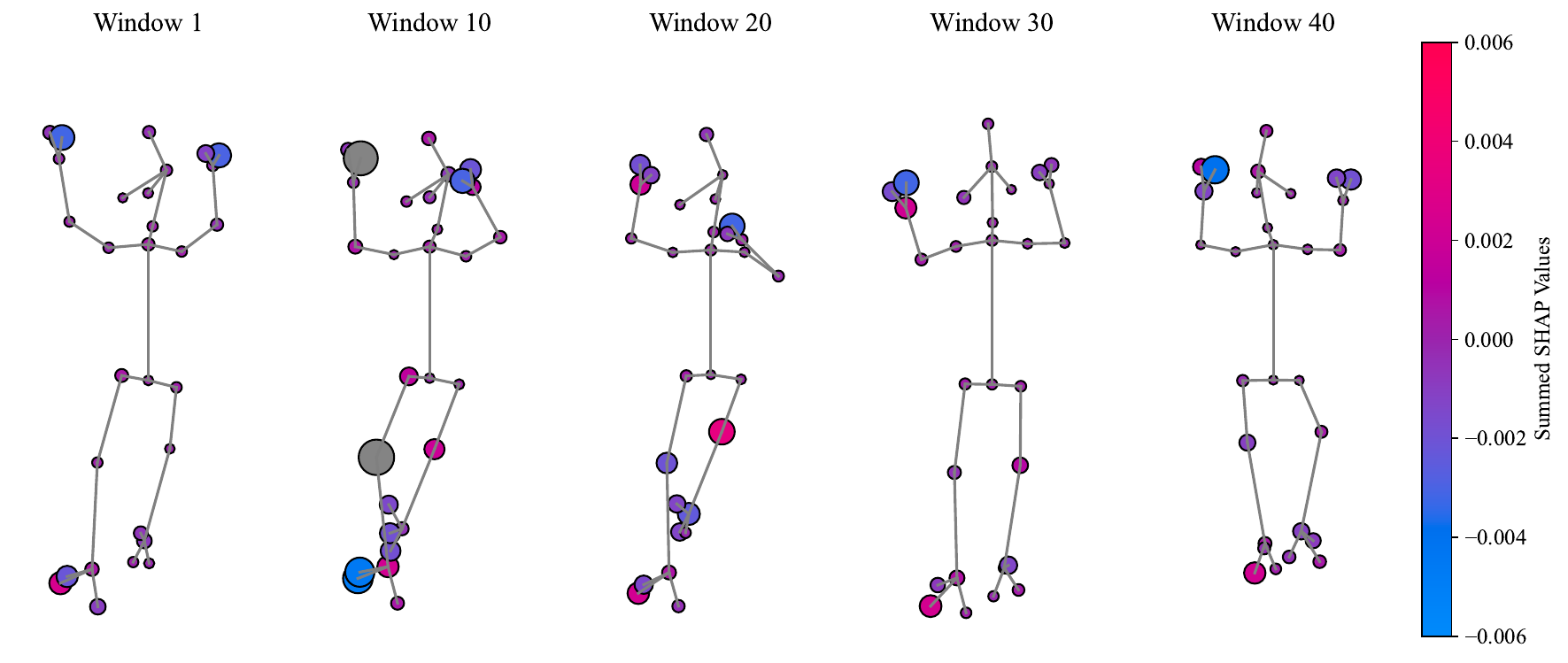}
        \caption{SHAP values of the velocity feature.}
        \label{fig:shap_window_vel}
    \end{subfigure}

    \begin{subfigure}[b]{0.67\textwidth}
        \centering
        \includegraphics[width=\textwidth]{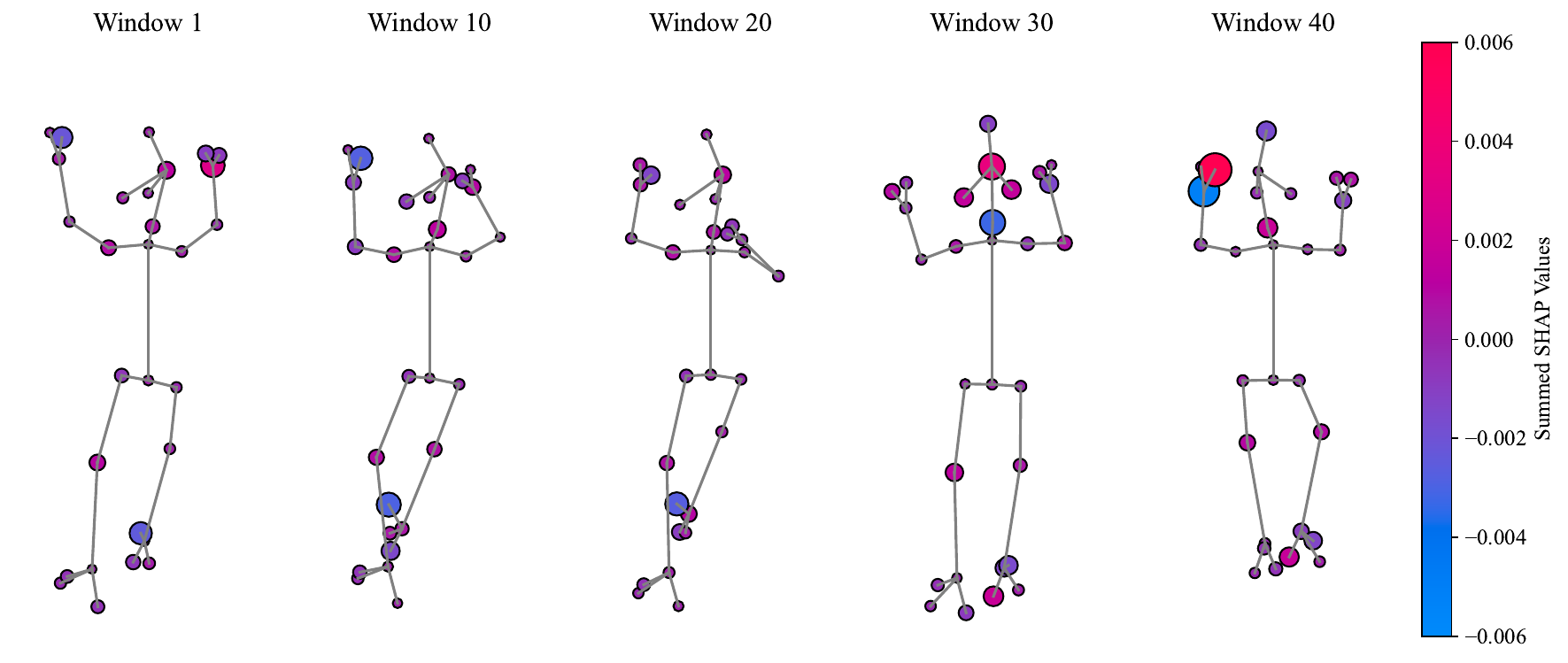}
        \caption{SHAP values of the bones feature.}
        \label{fig:shap_window_curv}
    \end{subfigure}

    \begin{subfigure}[b]{0.67\textwidth}
        \centering
        \includegraphics[width=\textwidth]{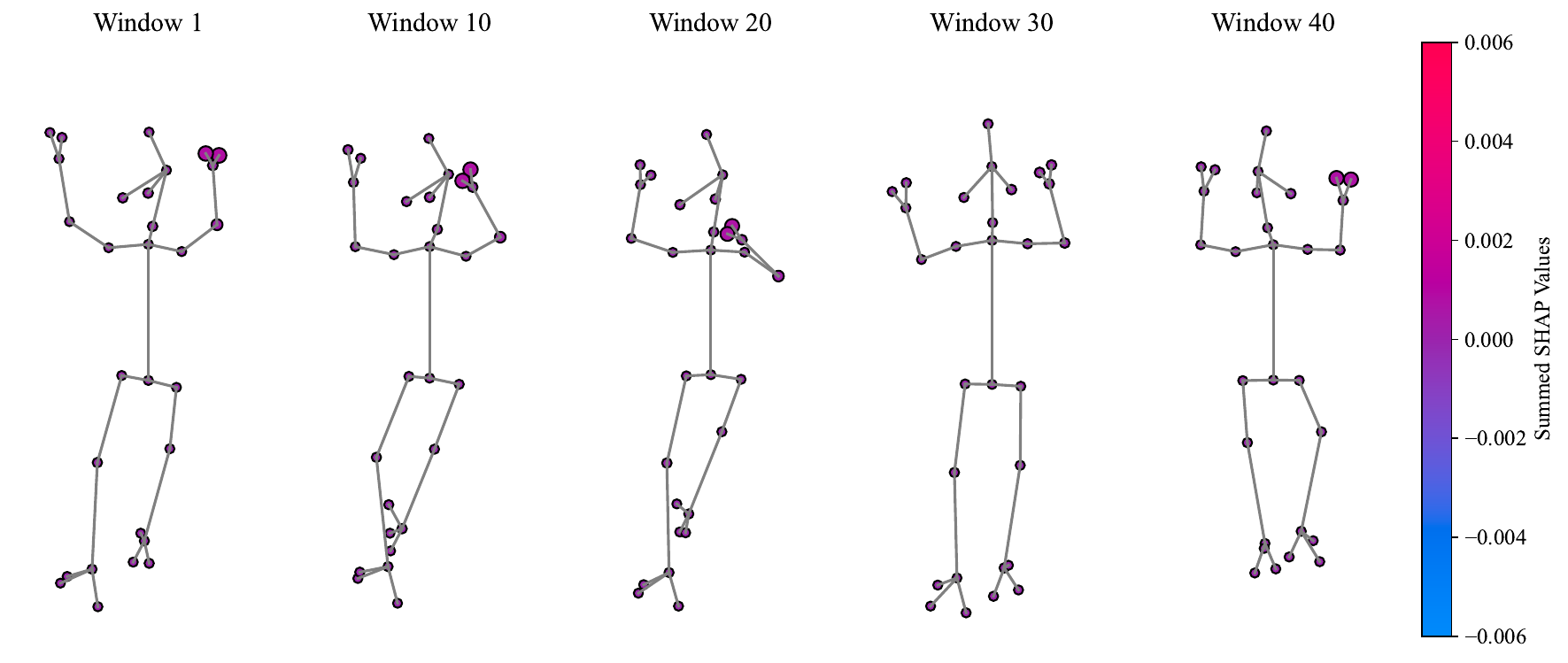}
        \caption{SHAP values of the joint feature.}
        \label{fig:shap_window_jerk}
    \end{subfigure}
    
    \caption{Comparison of the local SHAP values for the different input features.}
    \label{fig:shap_window_comparison}
\end{figure}

In Fig.~\ref{fig:shap_window_comparison}, the individual SHAP values for the input features $(J, V, B,$ and $A)$ are shown.
These plots illustrate the contribution of each feature to the model's predictions on a specific body key point. 
The variation in SHAP values across body key points, input features, and windows highlights their differing levels of importance, allowing for a detailed comparison of their individual impact on the prediction.

\section{NTU RGB+D Threshold Experiments}
\label{app:ntu_thresh}
This section presents the experimental results to show the performance behavior for the NTU RGB+D dataset under different edge matrix threshold perturbations.
The results of our experiments on classes 6, 11, and 16 are presented in Tables \ref{tab:ntu_class6}, \ref{tab:ntu_class11}, and \ref{tab:ntu_class16}, respectively.
Unlike the CP dataset, the NTU RGB+D dataset is especially sensitive to the masking perturbation, in which the corresponding edge $e_n$ in $\mathbf{E}$ is set to $1 \times 10^{-5}$, which results in a great performance degradation already after one perturbed joint.
Therefore, we conduct experiments in which the corresponding edge is perturbed with a percental threshold away from its original value.

\begin{table}[h]
    \centering
    \caption{\textbf{Performance metrics of class 6 (\textit{pick up}) with informed perturbation scenarios with varying percentual threshold settings.} 
    The table presents the accuracy of the important body key points (Acc. imp.), the accuracy of the unimportant body key points (Acc. $\neg$ imp.), PGI, and PGU for different perturbation percentages.
    }
    \label{tab:ntu_class6}
    \begin{minipage}{0.475\linewidth}
        \centering
        \resizebox{\linewidth}{!}{
        \begin{tabular}{llllll}
        \toprule
        
        Thresh. & Perturb. key points & Acc. imp. & Acc. $\neg$ imp. & PGI $(\uparrow)$ & PGU $(\downarrow)$ \\
        (\%) & (\#) &&&& \\
        \midrule
        5 & 1 &  0.975 & 0.972 & 0.004 & 0.001 \\
        5 & 2 &  0.972 & 0.972 & 0.001 & 0.001 \\
        5 & 3 &  0.975 & 0.972 & 0.004 & 0.001 \\
        5 & 4 &  0.972 & 0.972 & 0.001 & 0.001 \\
        5 & 5 &  0.972 & 0.972 & 0.001 & 0.001 \\
        5 & 6 &  0.972 & 0.972 & 0.001 & 0.001 \\
        5 & 7 &  0.972 & 0.975 & 0.001 & 0.004 \\
        5 & 8 &  0.972 & 0.972 & 0.001 & 0.001 \\
        5 & 9 &  0.972 & 0.972 & 0.001 & 0.001 \\
        5 & 10 &  0.972 & 0.972 & 0.001 & 0.001 \\
        \midrule
        10 & 1 &  0.968 & 0.972 & 0.003 & 0.001 \\
        10 & 2 &  0.972 & 0.978 & 0.001 & 0.007 \\
        10 & 3 &  0.972 & 0.978 & 0.001 & 0.007 \\
        10 & 4 &  0.972 & 0.975 & 0.001 & 0.004 \\
        10 & 5 &  0.972 & 0.975 & 0.001 & 0.004 \\
        10 & 6 &  0.972 & 0.972 & 0.001 & 0.001 \\
        10 & 7 &  0.981 & 0.968 & 0.010 & 0.003 \\
        10 & 8 &  0.975 & 0.972 & 0.004 & 0.001 \\
        10 & 9 &  0.975 & 0.975 & 0.004 & 0.004 \\
        10 & 10 &  0.972 & 0.975 & 0.001 & 0.004 \\
        \midrule
        15 & 1 &  0.965 & 0.975 & 0.006 & 0.004 \\
        15 & 2 &  0.962 & 0.978 & 0.009 & 0.007 \\
        15 & 3 &  0.965 & 0.975 & 0.006 & 0.004 \\
        15 & 4 &  0.968 & 0.975 & 0.003 & 0.004 \\
        15 & 5 &  0.965 & 0.972 & 0.006 & 0.001 \\
        15 & 6 &  0.972 & 0.968 & 0.001 & 0.003 \\
        15 & 7 &  0.972 & 0.968 & 0.001 & 0.003 \\
        15 & 8 &  0.975 & 0.972 & 0.004 & 0.001 \\
        15 & 9 &  0.975 & 0.972 & 0.004 & 0.001 \\
        15 & 10 &  0.968 & 0.972 & 0.003 & 0.001 \\
        \midrule
        20 & 1 &  0.956 & 0.975 & 0.015 & 0.004 \\
        20 & 2 &  0.959 & 0.965 & 0.012 & 0.006 \\
        20 & 3 &  0.959 & 0.968 & 0.012 & 0.003 \\
        20 & 4 &  0.968 & 0.968 & 0.003 & 0.003 \\
        20 & 5 &  0.959 & 0.965 & 0.012 & 0.006 \\
        20 & 6 &  0.972 & 0.959 & 0.001 & 0.012 \\
        20 & 7 &  0.959 & 0.959 & 0.012 & 0.012 \\
        20 & 8 &  0.965 & 0.956 & 0.006 & 0.015 \\
        20 & 9 &  0.968 & 0.956 & 0.003 & 0.015 \\
        20 & 10 &  0.962 & 0.959 & 0.009 & 0.012 \\
        \midrule
        25 & 1 &  0.949 & 0.972 & 0.022 & 0.001 \\
        25 & 2 &  0.943 & 0.956 & 0.028 & 0.015 \\
        25 & 3 &  0.927 & 0.956 & 0.044 & 0.015 \\
        25 & 4 &  0.959 & 0.940 & 0.012 & 0.031 \\
        25 & 5 &  0.946 & 0.930 & 0.025 & 0.041 \\
        25 & 6 &  0.956 & 0.934 & 0.015 & 0.037 \\
        25 & 7 &  0.956 & 0.934 & 0.015 & 0.037 \\
        25 & 8 &  0.953 & 0.934 & 0.018 & 0.037 \\
        25 & 9 &  0.953 & 0.934 & 0.018 & 0.037 \\
        25 & 10 &  0.934 & 0.930 & 0.037 & 0.041 \\
        \midrule
        \end{tabular}
        }
    \end{minipage}%
    \hfill
    \begin{minipage}{0.475\linewidth}
        \centering
        \resizebox{\linewidth}{!}{
        \begin{tabular}{llllll}
        
        \toprule
        Thresh. & Perturb. key points & Acc. imp. & Acc. $\neg$ imp. & PGI $(\uparrow)$ & PGU $(\downarrow)$ \\
        (\%) & (\#) &&&& \\
        \midrule
        30 & 1 &  0.924 & 0.965 & 0.047 & 0.006 \\
        30 & 2 &  0.908 & 0.930 & 0.063 & 0.041 \\
        30 & 3 &  0.902 & 0.911 & 0.069 & 0.060 \\
        30 & 4 &  0.886 & 0.886 & 0.085 & 0.085 \\
        30 & 5 &  0.858 & 0.883 & 0.113 & 0.088 \\
        30 & 6 &  0.848 & 0.873 & 0.123 & 0.098 \\
        30 & 7 &  0.848 & 0.867 & 0.123 & 0.104 \\
        30 & 8 &  0.851 & 0.864 & 0.120 & 0.107 \\
        30 & 9 &  0.848 & 0.858 & 0.123 & 0.113 \\
        30 & 10 &  0.835 & 0.854 & 0.136 & 0.117 \\
        \midrule
        35 & 1 &  0.883 & 0.940 & 0.088 & 0.031 \\
        35 & 2 &  0.851 & 0.918 & 0.120 & 0.053 \\
        35 & 3 &  0.759 & 0.883 & 0.212 & 0.088 \\
        35 & 4 &  0.617 & 0.839 & 0.354 & 0.132 \\
        35 & 5 &  0.560 & 0.807 & 0.411 & 0.164 \\
        35 & 6 &  0.541 & 0.782 & 0.430 & 0.189 \\
        35 & 7 &  0.566 & 0.772 & 0.405 & 0.199 \\
        35 & 8 &  0.573 & 0.763 & 0.398 & 0.208 \\
        35 & 9 &  0.557 & 0.753 & 0.414 & 0.218 \\
        35 & 10 &  0.544 & 0.734 & 0.427 & 0.237 \\
        \midrule
        40 & 1 &  0.775 & 0.930 & 0.196 & 0.041 \\
        40 & 2 &  0.766 & 0.886 & 0.205 & 0.085 \\
        40 & 3 &  0.646 & 0.826 & 0.325 & 0.145 \\
        40 & 4 &  0.424 & 0.763 & 0.547 & 0.208 \\
        40 & 5 &  0.244 & 0.722 & 0.727 & 0.249 \\
        40 & 6 &  0.177 & 0.699 & 0.794 & 0.272 \\
        40 & 7 &  0.212 & 0.696 & 0.759 & 0.275 \\
        40 & 8 &  0.266 & 0.671 & 0.705 & 0.300 \\
        40 & 9 &  0.228 & 0.674 & 0.743 & 0.297 \\
        40 & 10 &  0.203 & 0.668 & 0.768 & 0.303 \\
        \midrule
        45 & 1 &  0.655 & 0.899 & 0.316 & 0.072 \\
        45 & 2 &  0.611 & 0.813 & 0.360 & 0.158 \\
        45 & 3 &  0.516 & 0.753 & 0.455 & 0.218 \\
        45 & 4 &  0.361 & 0.677 & 0.610 & 0.294 \\
        45 & 5 &  0.165 & 0.630 & 0.806 & 0.341 \\
        45 & 6 &  0.063 & 0.608 & 0.908 & 0.363 \\
        45 & 7 &  0.051 & 0.598 & 0.920 & 0.373 \\
        45 & 8 &  0.089 & 0.592 & 0.882 & 0.379 \\
        45 & 9 &  0.082 & 0.595 & 0.889 & 0.376 \\
        45 & 10 &  0.079 & 0.589 & 0.892 & 0.382 \\
        \bottomrule
        \end{tabular}
        }
    \end{minipage}
\end{table}

\begin{table}[h]
    \centering
    \caption{\textbf{Performance metrics of class 11 (\textit{reading}) with informed perturbation scenarios with varying percentual threshold settings.} 
    The table presents the accuracy of the important body key points (Acc. imp.), the accuracy of the unimportant body key points (Acc. $\neg$ imp.), PGI, and PGU for different perturbation percentages.
    }
    \label{tab:ntu_class11}
    \begin{minipage}{0.45\linewidth}
        \centering
        \resizebox{\linewidth}{!}{
        \begin{tabular}{llllll}
        \toprule
        Thresh. & Perturb. key points & Acc. imp. & Acc. $\neg$ imp. & PGI $(\uparrow)$ & PGU $(\downarrow)$ \\
        (\%) & (\#) &&&& \\
        \midrule
        5 & 1 &  0.806 & 0.800 & 0.009 & 0.003 \\
        5 & 2 &  0.816 & 0.800 & 0.019 & 0.003 \\
        5 & 3 &  0.819 & 0.797 & 0.022 & 0.000 \\
        5 & 4 &  0.816 & 0.797 & 0.019 & 0.000 \\
        5 & 5 &  0.813 & 0.797 & 0.016 & 0.000 \\
        5 & 6 &  0.825 & 0.784 & 0.028 & 0.013 \\
        5 & 7 &  0.822 & 0.787 & 0.025 & 0.010 \\
        5 & 8 &  0.819 & 0.784 & 0.022 & 0.013 \\
        5 & 9 &  0.813 & 0.784 & 0.016 & 0.013 \\
        5 & 10 &  0.822 & 0.784 & 0.025 & 0.013 \\
        \midrule
        10 & 1 &  0.819 & 0.797 & 0.022 & 0.000 \\
        10 & 2 &  0.825 & 0.803 & 0.028 & 0.006 \\
        10 & 3 &  0.838 & 0.800 & 0.041 & 0.003 \\
        10 & 4 &  0.851 & 0.797 & 0.054 & 0.000 \\
        10 & 5 &  0.851 & 0.790 & 0.054 & 0.007 \\
        10 & 6 &  0.854 & 0.781 & 0.057 & 0.016 \\
        10 & 7 &  0.854 & 0.781 & 0.057 & 0.016 \\
        10 & 8 &  0.854 & 0.784 & 0.057 & 0.013 \\
        10 & 9 &  0.860 & 0.784 & 0.063 & 0.013 \\
        10 & 10 &  0.863 & 0.784 & 0.066 & 0.013 \\
        \midrule
        15 & 1 &  0.825 & 0.797 & 0.028 & 0.000 \\
        15 & 2 &  0.848 & 0.787 & 0.051 & 0.010 \\
        15 & 3 &  0.867 & 0.787 & 0.070 & 0.010 \\
        15 & 4 &  0.879 & 0.790 & 0.082 & 0.007 \\
        15 & 5 &  0.876 & 0.781 & 0.079 & 0.016 \\
        15 & 6 &  0.876 & 0.778 & 0.079 & 0.019 \\
        15 & 7 &  0.873 & 0.778 & 0.076 & 0.019 \\
        15 & 8 &  0.883 & 0.784 & 0.086 & 0.013 \\
        15 & 9 &  0.879 & 0.781 & 0.082 & 0.016 \\
        15 & 10 &  0.876 & 0.781 & 0.079 & 0.016 \\
        \midrule
        20 & 1 &  0.829 & 0.787 & 0.032 & 0.010 \\
        20 & 2 &  0.848 & 0.778 & 0.051 & 0.019 \\
        20 & 3 &  0.870 & 0.765 & 0.073 & 0.032 \\
        20 & 4 &  0.883 & 0.775 & 0.086 & 0.022 \\
        20 & 5 &  0.886 & 0.759 & 0.089 & 0.038 \\
        20 & 6 &  0.870 & 0.762 & 0.073 & 0.035 \\
        20 & 7 &  0.854 & 0.752 & 0.057 & 0.045 \\
        20 & 8 &  0.844 & 0.756 & 0.047 & 0.041 \\
        20 & 9 &  0.854 & 0.756 & 0.057 & 0.041 \\
        20 & 10 &  0.854 & 0.752 & 0.057 & 0.045 \\
        \midrule
        25 & 1 &  0.790 & 0.778 & 0.007 & 0.019 \\
        25 & 2 &  0.810 & 0.756 & 0.013 & 0.041 \\
        25 & 3 &  0.813 & 0.746 & 0.016 & 0.051 \\
        25 & 4 &  0.819 & 0.752 & 0.022 & 0.045 \\
        25 & 5 &  0.816 & 0.740 & 0.019 & 0.057 \\
        25 & 6 &  0.794 & 0.727 & 0.003 & 0.070 \\
        25 & 7 &  0.781 & 0.714 & 0.016 & 0.083 \\
        25 & 8 &  0.759 & 0.708 & 0.038 & 0.089 \\
        25 & 9 &  0.721 & 0.692 & 0.076 & 0.105 \\
        25 & 10 &  0.670 & 0.689 & 0.127 & 0.108 \\
        \midrule
        \end{tabular}
        }
    \end{minipage}%
    \hfill
    \begin{minipage}{0.45\linewidth}
        \centering
        \resizebox{\linewidth}{!}{
        \begin{tabular}{llllll}
        
        \toprule
        Thresh. & Perturb. key points & Acc. imp. & Acc. $\neg$ imp. & PGI $(\uparrow)$ & PGU $(\downarrow)$ \\
        (\%) & (\#) &&&& \\
        \midrule
        30 & 1 &  0.727 & 0.771 & 0.07 & 0.026 \\
        30 & 2 &  0.746 & 0.724 & 0.051 & 0.073 \\
        30 & 3 &  0.714 & 0.702 & 0.083 & 0.095 \\
        30 & 4 &  0.683 & 0.663 & 0.114 & 0.134 \\
        30 & 5 &  0.629 & 0.625 & 0.168 & 0.172 \\
        30 & 6 &  0.594 & 0.606 & 0.203 & 0.191 \\
        30 & 7 &  0.549 & 0.578 & 0.248 & 0.219 \\
        30 & 8 &  0.489 & 0.565 & 0.308 & 0.232 \\
        30 & 9 &  0.435 & 0.556 & 0.362 & 0.241 \\
        30 & 10 &  0.362 & 0.546 & 0.435 & 0.251 \\
        \midrule
        35 & 1 &  0.597 & 0.733 & 0.200 & 0.064 \\
        35 & 2 &  0.568 & 0.667 & 0.229 & 0.130 \\
        35 & 3 &  0.505 & 0.632 & 0.292 & 0.165 \\
        35 & 4 &  0.470 & 0.568 & 0.327 & 0.229 \\
        35 & 5 &  0.371 & 0.486 & 0.426 & 0.311 \\
        35 & 6 &  0.305 & 0.397 & 0.492 & 0.400 \\
        35 & 7 &  0.232 & 0.368 & 0.565 & 0.429 \\
        35 & 8 &  0.187 & 0.337 & 0.610 & 0.460 \\
        35 & 9 &  0.117 & 0.321 & 0.680 & 0.476 \\
        35 & 10 &  0.086 & 0.314 & 0.711 & 0.483 \\
        \midrule
        40 & 1 &  0.432 & 0.695 & 0.365 & 0.102 \\
        40 & 2 &  0.349 & 0.600 & 0.448 & 0.197 \\
        40 & 3 &  0.235 & 0.495 & 0.562 & 0.302 \\
        40 & 4 &  0.178 & 0.365 & 0.619 & 0.432 \\
        40 & 5 &  0.124 & 0.302 & 0.673 & 0.495 \\
        40 & 6 &  0.095 & 0.251 & 0.702 & 0.546 \\
        40 & 7 &  0.076 & 0.206 & 0.721 & 0.591 \\
        40 & 8 &  0.041 & 0.190 & 0.756 & 0.607 \\
        40 & 9 &  0.016 & 0.184 & 0.781 & 0.613 \\
        40 & 10 &  0.013 & 0.181 & 0.784 & 0.616 \\
        \midrule
        45 & 1 &  0.333 & 0.673 & 0.464 & 0.124 \\
        45 & 2 &  0.184 & 0.505 & 0.613 & 0.292 \\
        45 & 3 &  0.117 & 0.346 & 0.680 & 0.451 \\
        45 & 4 &  0.089 & 0.232 & 0.708 & 0.565 \\
        45 & 5 &  0.035 & 0.175 & 0.762 & 0.622 \\
        45 & 6 &  0.029 & 0.140 & 0.768 & 0.657 \\
        45 & 7 &  0.022 & 0.127 & 0.775 & 0.670 \\
        45 & 8 &  0.006 & 0.111 & 0.791 & 0.686 \\
        45 & 9 &  0.003 & 0.111 & 0.794 & 0.686 \\
        45 & 10 &  0.000 & 0.111 & 0.797 & 0.686 \\
        \bottomrule
        \end{tabular}
        }
    \end{minipage}
\end{table}

\begin{table}[h]
    \centering
    \caption{\textbf{Performance metrics of class 16 (\textit{put on a shoe}) with informed perturbation scenarios with varying percentual threshold settings.} 
    The table presents the accuracy of the important body key points (Acc. imp.), the accuracy of the unimportant body key points (Acc. $\neg$ imp.), PGI, and PGU for different perturbation percentages.
    }
    \label{tab:ntu_class16}
    \begin{minipage}{0.45\linewidth}
        \centering
        \resizebox{\linewidth}{!}{
        \begin{tabular}{llllll}
        \toprule
        Thresh. & Perturb. key points & Acc. imp. & Acc. $\neg$ imp. & PGI $(\uparrow)$ & PGU $(\downarrow)$ \\
        (\%) & (\#) &&&& \\
        \midrule
        5 & 1 &  0.838 & 0.841 & 0.007 & 0.004 \\
        5 & 2 &  0.838 & 0.841 & 0.007 & 0.004 \\
        5 & 3 &  0.848 & 0.841 & 0.003 & 0.004 \\
        5 & 4 &  0.844 & 0.841 & 0.001 & 0.004 \\
        5 & 5 &  0.841 & 0.841 & 0.004 & 0.004 \\
        5 & 6 &  0.844 & 0.841 & 0.001 & 0.004 \\
        5 & 7 &  0.838 & 0.841 & 0.007 & 0.004 \\
        5 & 8 &  0.841 & 0.841 & 0.004 & 0.004 \\
        5 & 9 &  0.854 & 0.841 & 0.009 & 0.004 \\
        5 & 10 &  0.848 & 0.841 & 0.003 & 0.004 \\
        \midrule
        10 & 1 &  0.829 & 0.838 & 0.016 & 0.007 \\
        10 & 2 &  0.838 & 0.829 & 0.007 & 0.016 \\
        10 & 3 &  0.841 & 0.841 & 0.004 & 0.004 \\
        10 & 4 &  0.841 & 0.844 & 0.004 & 0.001 \\
        10 & 5 &  0.832 & 0.844 & 0.013 & 0.001 \\
        10 & 6 &  0.841 & 0.841 & 0.004 & 0.004 \\
        10 & 7 &  0.838 & 0.835 & 0.007 & 0.010 \\
        10 & 8 &  0.838 & 0.835 & 0.007 & 0.010 \\
        10 & 9 &  0.844 & 0.835 & 0.001 & 0.010 \\
        10 & 10 &  0.841 & 0.835 & 0.004 & 0.010 \\
        \midrule
        15 & 1 &  0.813 & 0.835 & 0.032 & 0.010 \\
        15 & 2 &  0.810 & 0.816 & 0.035 & 0.029 \\
        15 & 3 &  0.825 & 0.819 & 0.020 & 0.026 \\
        15 & 4 &  0.819 & 0.819 & 0.026 & 0.026 \\
        15 & 5 &  0.822 & 0.822 & 0.023 & 0.023 \\
        15 & 6 &  0.822 & 0.819 & 0.023 & 0.026 \\
        15 & 7 &  0.819 & 0.819 & 0.026 & 0.026 \\
        15 & 8 &  0.829 & 0.819 & 0.016 & 0.026 \\
        15 & 9 &  0.825 & 0.822 & 0.020 & 0.023 \\
        15 & 10 &  0.813 & 0.822 & 0.032 & 0.023 \\
        \midrule
        20 & 1 &  0.771 & 0.835 & 0.074 & 0.010 \\
        20 & 2 &  0.775 & 0.803 & 0.070 & 0.042 \\
        20 & 3 &  0.756 & 0.806 & 0.089 & 0.039 \\
        20 & 4 &  0.759 & 0.813 & 0.086 & 0.032 \\
        20 & 5 &  0.775 & 0.806 & 0.070 & 0.039 \\
        20 & 6 &  0.787 & 0.797 & 0.058 & 0.048 \\
        20 & 7 &  0.778 & 0.790 & 0.067 & 0.055 \\
        20 & 8 &  0.759 & 0.781 & 0.086 & 0.064 \\
        20 & 9 &  0.762 & 0.787 & 0.083 & 0.058 \\
        20 & 10 &  0.746 & 0.787 & 0.099 & 0.058 \\
        \midrule
        25 & 1 &  0.698 & 0.816 & 0.147 & 0.029 \\
        25 & 2 &  0.721 & 0.800 & 0.124 & 0.045 \\
        25 & 3 &  0.717 & 0.787 & 0.128 & 0.058 \\
        25 & 4 &  0.721 & 0.768 & 0.124 & 0.077 \\
        25 & 5 &  0.733 & 0.756 & 0.112 & 0.089 \\
        25 & 6 &  0.714 & 0.743 & 0.131 & 0.102 \\
        25 & 7 &  0.711 & 0.749 & 0.134 & 0.096 \\
        25 & 8 &  0.721 & 0.743 & 0.124 & 0.102 \\
        25 & 9 &  0.708 & 0.740 & 0.137 & 0.105 \\
        25 & 10 &  0.686 & 0.743 & 0.159 & 0.102 \\
        \midrule
        \end{tabular}
        }
    \end{minipage}%
    \hfill
    \begin{minipage}{0.45\linewidth}
        \centering
        \resizebox{\linewidth}{!}{
        \begin{tabular}{llllll}
        
        \toprule
        Thresh. & Perturb. key points & Acc. imp. & Acc. $\neg$ imp. & PGI $(\uparrow)$ & PGU $(\downarrow)$ \\
        (\%) & (\#) &&&& \\
        \midrule
        30 & 1 &  0.635 & 0.810 & 0.210 & 0.035 \\
        30 & 2 &  0.638 & 0.775 & 0.207 & 0.070 \\
        30 & 3 &  0.629 & 0.737 & 0.216 & 0.108 \\
        30 & 4 &  0.644 & 0.692 & 0.201 & 0.153 \\
        30 & 5 &  0.606 & 0.692 & 0.239 & 0.153 \\
        30 & 6 &  0.610 & 0.667 & 0.235 & 0.178 \\
        30 & 7 &  0.603 & 0.641 & 0.242 & 0.204 \\
        30 & 8 &  0.600 & 0.641 & 0.245 & 0.204 \\
        30 & 9 &  0.622 & 0.641 & 0.223 & 0.204 \\
        30 & 10 &  0.587 & 0.629 & 0.258 & 0.216 \\
        \midrule
        35 & 1 &  0.533 & 0.784 & 0.312 & 0.061 \\
        35 & 2 &  0.444 & 0.727 & 0.401 & 0.118 \\
        35 & 3 &  0.406 & 0.686 & 0.439 & 0.159 \\
        35 & 4 &  0.397 & 0.629 & 0.448 & 0.216 \\
        35 & 5 &  0.400 & 0.575 & 0.445 & 0.270\\
        35 & 6 &  0.429 & 0.540 & 0.416 & 0.305 \\
        35 & 7 &  0.406 & 0.521 & 0.439 & 0.324 \\
        35 & 8 &  0.422 & 0.511 & 0.423 & 0.334 \\
        35 & 9 &  0.432 & 0.505 & 0.413 & 0.340 \\
        35 & 10 &  0.416 & 0.502 & 0.429 & 0.343 \\
        \midrule
        40 & 1 &  0.422 & 0.756 & 0.423 & 0.089 \\
        40 & 2 &  0.324 & 0.670 & 0.521 & 0.175 \\
        40 & 3 &  0.260 & 0.600 & 0.585 & 0.245 \\
        40 & 4 &  0.260 & 0.521 & 0.585 & 0.324 \\
        40 & 5 &  0.248 & 0.473 & 0.597 & 0.372 \\
        40 & 6 &  0.273 & 0.438 & 0.572 & 0.407 \\
        40 & 7 &  0.267 & 0.416 & 0.578 & 0.429 \\
        40 & 8 &  0.254 & 0.397 & 0.591 & 0.448 \\
        40 & 9 &  0.302 & 0.397 & 0.543 & 0.448 \\
        40 & 10 &  0.270 & 0.394 & 0.575 & 0.451 \\
        \midrule
        45 & 1 &  0.333 & 0.714 & 0.512 & 0.131 \\
        45 & 2 &  0.283 & 0.597 & 0.562 & 0.248 \\
        45 & 3 &  0.225 & 0.492 & 0.620 & 0.353 \\
        45 & 4 &  0.175 & 0.387 & 0.670 & 0.458 \\
        45 & 5 &  0.165 & 0.362 & 0.680 & 0.483 \\
        45 & 6 &  0.143 & 0.324 & 0.702 & 0.521 \\
        45 & 7 &  0.152 & 0.308 & 0.693 & 0.537 \\
        45 & 8 &  0.162 & 0.286 & 0.683 & 0.559 \\
        45 & 9 &  0.162 & 0.283 & 0.683 & 0.562 \\
        45 & 10 &  0.156 & 0.283 & 0.689 & 0.562 \\
        \bottomrule
        \end{tabular}
        }
    \end{minipage}
\end{table}

\end{document}